\documentclass[10pt,twocolumn,letterpaper]{article}

\usepackage[pagenumbers]{cvpr} 

\usepackage[dvipsnames]{xcolor}
\usepackage{colortbl}

\definecolor{placeholder}{rgb}{0.6,0.8,0.95}

\newcommand{\modelname}{\emph{MVG}-Splatting\xspace}

\definecolor{amber}{rgb}{1.0, 0.75, 0.0}

\definecolor{visible-blue}{rgb}{0.286, 0.525, 0.910}

\definecolor{tabfirst}{rgb}{1, 0.7, 0.7} 
\definecolor{tabsecond}{rgb}{1, 0.85, 0.7} 
\definecolor{tabthird}{rgb}{1, 1, 0.7} 
\usepackage{overpic}
\usepackage{wrapfig}
\usepackage{amsmath}
\usepackage{empheq}
\usepackage{comment}
\usepackage{makecell}
\usepackage{multirow}
\usepackage{multicol}
\usepackage{float}
\usepackage{marvosym}
\usepackage{algorithm}
\usepackage{algorithmicx}
\usepackage{algpseudocode}

\definecolor{cvprblue}{rgb}{0.21,0.49,0.74}
\usepackage[pagebackref,breaklinks,colorlinks,citecolor=cvprblue]{hyperref}

\def\paperID{****} 
\def\confName{CVPR}
\def\confYear{2024}

\title{MVG-Splatting: Multi-View Guided Gaussian Splatting with Adaptive Quantile-Based Geometric Consistency Densification}

\author{Zhuoxiao Li $^{1,2, *}$ \quad Shanliang Yao $^{1,2, *}$  \quad Yijie Chu $^{1,2}$ \quad \'{A}ngel F. Garc\'{i}a-Fern\'{a}ndez $^{1,3}$  \quad  Yong Yue $^{1,2}$ \\ 
Eng Gee Lim $^{1,2}$ \quad Xiaohui Zhu $^{1,2}$  \\
$^1$ University of Liverpool, $^2$ Xi'an Jiaotong-Liverpool University, \\$^3$ ARIES Research Centre, Universidad Antonio de Nebrija \\
}
\begin{document}

\twocolumn[{%
	\renewcommand
	\twocolumn[1][]{#1}%
	\maketitle
	\begin{center}
		\centering
		\vspace{-15pt}
            \includegraphics[width=\textwidth]{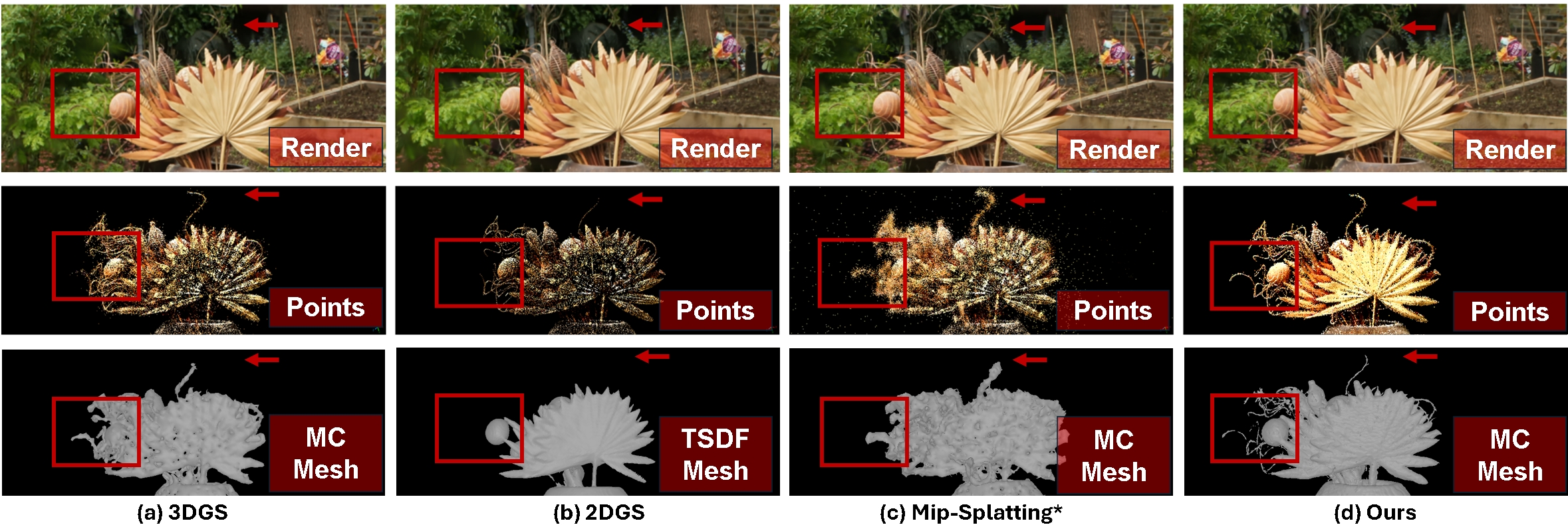}
		\captionof{figure}{\small
            \textbf{\modelname} represents the scene with denser Gaussian point clouds. We design an adaptive densification method guided by multi-view inputs for the GS-based capability to render depth, directing densification in areas that are under-reconstructed and require intensive reconstruction. From top to bottom: the rendered scene, the Gaussian point cloud, and the extracted mesh. Our method (d), through more uniform densification, can directly utilize the Marching Cubes (MC) method to extract detailed meshes.
		}
		\label{fig:teaser}
	\end{center}
}]
\maketitle

\begingroup
\renewcommand\thefootnote{}\footnotetext{* denotes equal contribution.}
\endgroup
\begin{abstract}
\small
\vspace{-5mm}
In the rapidly evolving field of 3D reconstruction, 3D Gaussian Splatting (3DGS) and 2D Gaussian Splatting (2DGS) represent significant advancements. Although 2DGS compresses 3D Gaussian primitives into 2D Gaussian surfels to effectively enhance mesh extraction quality, this compression can potentially lead to a decrease in rendering quality. Additionally, unreliable densification processes and the calculation of depth through the accumulation of opacity can compromise the detail of mesh extraction. To address this issue, we introduce MVG-Splatting, a solution guided by Multi-View considerations. Specifically, we integrate an optimized method for calculating normals, which, combined with image gradients, helps rectify inconsistencies in the original depth computations. Additionally, utilizing projection strategies akin to those in Multi-View Stereo (MVS), we propose an adaptive quantile-based method that dynamically determines the level of additional densification guided by depth maps, from coarse to fine detail. Experimental evidence demonstrates that our method not only resolves the issues of rendering quality degradation caused by depth discrepancies but also facilitates direct mesh extraction from dense Gaussian point clouds using the Marching Cubes algorithm. This approach significantly enhances the overall fidelity and accuracy of the 3D reconstruction process, ensuring that both the geometric details and visual quality. The project is available at \href{https://mvgsplatting.github.io/}{https://mvgsplatting.github.io/} .
\end{abstract}    
\section{Introduction}
\label{sec:intro}
\small
Multi-view image-based 3D reconstruction and rendering techniques have undergone transformative development, progressing from traditional Multi-view Stereo (MVS) methods \cite{furukawa2009accurate, bleyer2011patchmatch,xu2019multi,xu2022multi} to learned-based MVS approaches \cite{yao2018mvsnet,chen2019pointmvs,yu2020fastmvs,ding2022transmvsnet,su2023edgemvs,zhang2023geomvsnet,wu2024gomvs}, and advancing further into Neural Radiance Fields (NeRF) \cite{mildenhall2021nerf,zhang2020nerf++,xu2022pointnerf,barron2021mipnerf,barron2022mipnerf360}. One of the latest advancements is 3D Gaussian Splatting (3DGS) \cite{3dgs}, which has attracted considerable attention for its efficient computational model and superior processing speeds. Building on the 3DGS, 2DGS \cite{huang20242dgs} utilized 2D oriented Gaussian disks \cite{pfister2000surfels} to improve the limitation of 3DGS's geometric accuracy and stability in surface reconstruction. This method focused on enhancing the consistency of views across normals and depth maps and  adopting truncated signed distance function (TSDF) fusion \cite{werner2014tsdf} for better surface detailed mesh extraction after training. 

\begin{figure}[t]
  \centering
   \includegraphics[width=\linewidth]{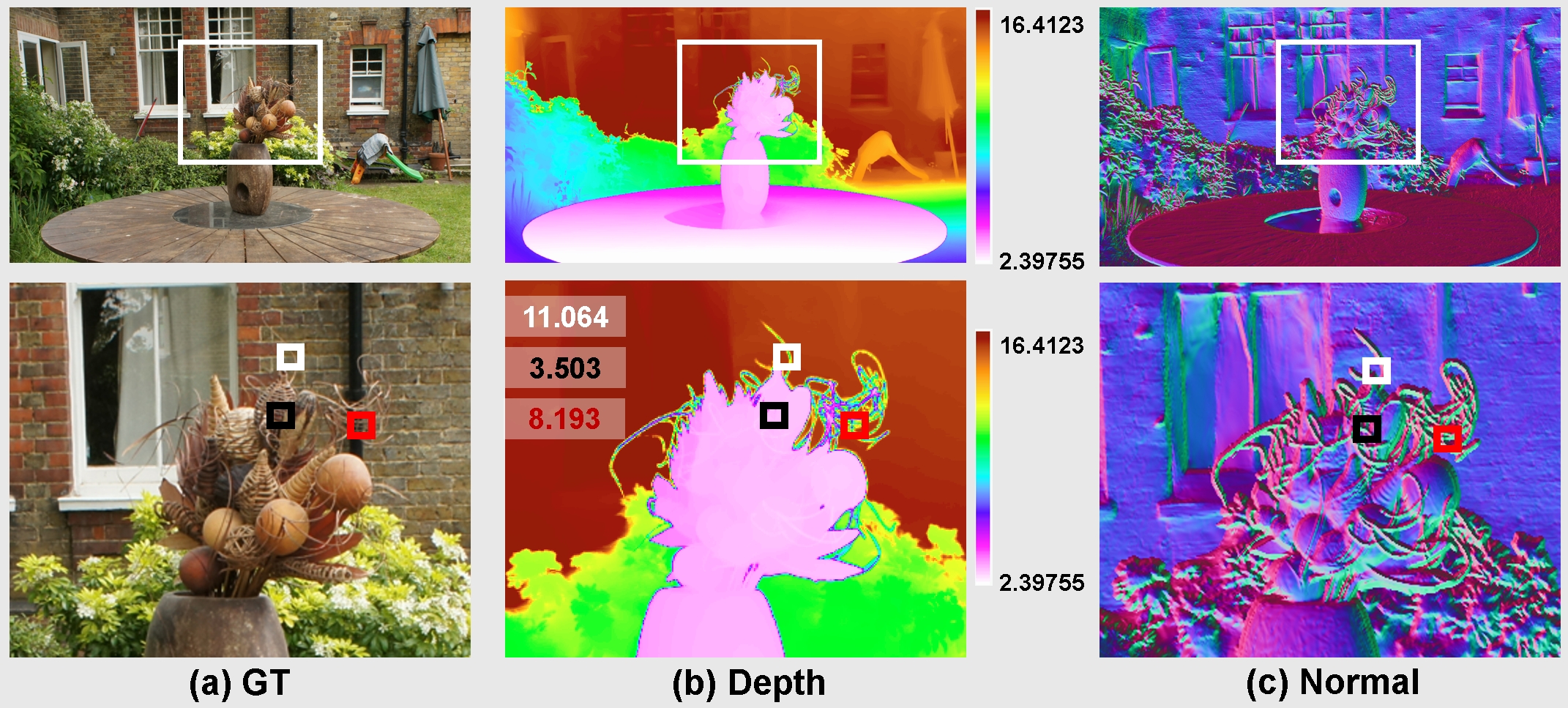}
   \caption{\textbf{Quantitative and Qualitative Visualizations of 2DGS's Depth and Normal.}  The three images represent the vase's Ground Truth image, surface depth map, and surface normal map, respectively. For a detailed quantitative analysis, depth values were extracted from three specific pixels on the vase, highlighted by boxes colored in \textit{white}, \textit{black}, and \textit{red}. The depth values obtained for these pixels are \textit{11.064}, \textit{3.503}, and \textit{8.913}, respectively.  }
   \label{fig:2dgs_depth_normal}
\end{figure}
In practice, we observe that the TSDF fusion-based GS methods \cite{huang20242dgs,zhang2024rade} for mesh reconstruction are capable of reconstructing objects with rich geometry. However, due to the degeneracy of a 2D Gaussian disk \cite{huang20242dgs}, it occasionally lacks fine details, such as tentacles, fur, and foliage. Figures \ref{fig:teaser}(b) and Figure \ref{fig:2dgs_depth_normal} illustrate a case study on the Mip-NeRF 360 dataset \cite{barron2022mipnerf360} employing 2DGS \cite{huang20242dgs}. It is evident that the depth computation \cite{huang20242dgs} results of 2DGS in Figure \ref{fig:2dgs_depth_normal}(b) are adversely affected by opacity in fine-grained geometric areas. Specifically, high opacity in background regions (building) impacts the depth computation of areas with low opacity, resulting in inaccuracies in the rendered depth. Consequently, during the TSDF reconstruction process, the voxels in these regions are not accurately represented, leading to errors in the geometric detail of the reconstructed surfaces. However, as shown in Figure \ref{fig:2dgs_depth_normal}(c), if these local variations in depth remain visually consistent (despite errors in the overall depth values) the computed normals can still effectively depict the true surface geometry and contours of the object.

Based on the analyses and observations above, it has been determined that TSDF-based GS studies \cite{huang20242dgs,zhang2024rade} struggle to accurately extract fine surface details. Despite efforts involving the use of smaller voxels for surface extraction, the inaccuracies inherent in depth representations ultimately hinder precise detail retrieval. To address these limitations, we propose a method that leverages depth-normal mutual optimization to guide a more refined densification process, thereby enriching the detail representation in rendering scenes and surface extraction. Specifically, building upon the 2DGS \cite{huang20242dgs} framework, we employ a more robust approach for recalculating surface normals. The recalculated normals, in conjunction with gradients from the original images, guide the refinement and accuracy of the rendered depth maps. Subsequently, we design an efficient multi-level densification method based on multi-view geometric consistency \cite{yao2018mvsnet}, which guides the refined depth maps in projecting onto under-reconstructed areas. Distinct from previous GS-based geometric reconstructions \cite{cheng2024gaussianpro,liu2024fast,wolf2024surface,chen2024mvsplat}, ours prior step yields high-quality and uniformly densified Gaussian point clouds, enabling us direct surface extraction using the Marching Cubes \cite{lorensen1998marching} method on the point cloud. In addition, we dynamically determine voxel sizes locally for each densified Gaussian point cloud and employ multi-view normal maps to smooth and optimize surface normals of extracted meshes, ultimately extracting high-detail surface information.

To demonstrate the efficacy of MVG-Splatting, we conducted experiments across three datasets \cite{barron2022mipnerf360, jensen2014dtu,knapitsch2017tanks}. The results indicate that MVG-Splatting is capable of extracting meshes rich in detail and achieves advanced results in novel view synthesis (NVS) rendering. Furthermore, we extended our evaluation to a large-scale drone aerial dataset \cite{UrbanScene3D} at low resolution to validate the performance of our method on extensive datasets. This broader testing confirms that MVG-Splatting not only excels in controlled environments but also scales effectively to handle large-scale data challenges, maintaining high fidelity in mesh detail and rendering quality.


\section{Related work}
\label{sec:Related_work}

\subsection{Novel View Synthesis} 
Neural Radiance Fields (NeRF) \cite{mildenhall2021nerf} synthesize novel views by casting rays through the scene and using volume rendering techniques to estimate color and density at different points along each ray, integrating these to produce pixel colors. The works on NeRFs are primarily focused on using more efficient network architectures \cite{zhang2020nerf++,mueller2022instantngp,fridovich2022plenoxels}, improved sampling strategies \cite{barron2021mipnerf,barron2022mipnerf360}, and hardware acceleration \cite{mueller2022instantngp} to improve rendering quality and decrease training time. Tailoring NeRF implementations to specific use cases is essential, whether it is for model reconstruction, where rapid training is crucial \cite{oechsle2021unisurf,wang2021neus,yariv2021volume,oechsle2021unisurf}, or for large-scale outdoor reconstruction, where accuracy is paramount \cite{tancik2022blocknerf,mi2023switchnerf}.

Recently, 3DGS \cite{3dgs} has significantly streamlined the rendering process by employing 3D Gaussian primitives in place of the traditional dense MLP-based sampling.  Subsequent research efforts have focused on further enhancements, achieving commendable progress in two main areas: improving the rendering quality of 3DGS \cite{yu2024mipsplatting,yan2024multi,lu2024scaffold,song2024sags} and enhancing the quality of mesh extraction \cite{huang20242dgs,zhang2024rade,chen2024pgsr,Yu2024GOF}. 
Among them, 2DGS \cite{huang20242dgs} extends the fast training capabilities of 3DGS by projecting 3D volumes onto 2D oriented Gaussian disks. This adaptation specifically aims to enhance the geometric accuracy of surface details, however, it decreases the rendering quality.

In this work, we further develop strategies that leverage multi-view geometric consistency to guide the densification process. This method effectively harnesses the consistency across different views to optimize the placement and density of Gaussian points in the reconstruction, thereby improving both the detail and the integrity of the rendered scenes. 

\subsection{3D Reconstruction}
In the domain of 3D reconstruction, significant progress has been made over the years, particularly through the development and refinement of techniques such as Multi-view Stereo (MVS) \cite{wu2024gomvs,su2023edgemvs,yu2020fastmvs,yao2018mvsnet,ding2022transmvsnet} and Signed Distance Functions (SDF) \cite{werner2014tsdf}. MVS relies heavily on feature matching and stereo correspondence to estimate depth, making it highly effective for capturing complex surfaces and textures in natural scenes. Innovations such as semi-global matching \cite{bleyer2011patchmatch} and the integration of machine learning \cite{yao2018mvsnet,gu2020cascade} have greatly enhanced its accuracy and efficiency.  SDF represents a surface by a function that gives the shortest distance to the surface boundary. It have been incorporated into deep learning frameworks to create differentiable rendering pipelines that support tasks like shape optimization and inverse rendering \cite{yariv2023bakedsdf,wang2021neus,yariv2021volume,li2023neuralangelo}. However, MVS struggles with texture-poor environments and high computational demands and SDF face challenges in handling large scenes and complex topology changes. In contrast, our method re-optimizes the rendering depth of 2DGS \cite{huang20242dgs}, utilizing normals combined with image gradients to guide a more enriched and accurate method for the projection \cite{yao2018mvsnet} and initialization of Gaussian primitives \cite{cheng2024gaussianpro}. Additionally, it employs an adaptive Marching Cubes \cite{lorensen1998marching} method integrated with TSDF \cite{werner2014tsdf} to efficiently extract meshes rich in detail.

\subsection{3DGS-based Mesh Extraction}
Inspired by the efficient and high-fidelity performance of 3DGS \cite{3dgs}, researchers have harnessed its inherent explicit rendering properties to directly extract refined surfaces from optimized Gaussian primitives. Notably, the use of Poisson surface reconstruction \cite{guedon2023sugar} and TSDF fusion \cite{huang20242dgs,zhang2024rade,chen2024pgsr} stand out as two of the most representative methods in this domain. Additionally, several studies have focused on optimizing normal calculation \cite{chen2024pgsr,zhang2024rade} methods or employing opacity fields to enhance rendering \cite{Yu2024GOF}. Among these, Gaussian Opacity Fields (GOF) directly extract surfaces within opacity fields and utilize innovative tetrahedral-based mesh extraction techniques to achieve astonishing mesh detail \cite{Yu2024GOF}. However, compared to other methods, GOF significantly increases both training and mesh extraction times.

In our work, owing to the acquisition of a denser point cloud compared to other GS-based approaches \cite{3dgs,yu2024mipsplatting,huang20242dgs,Yu2024GOF}, we directly perform Marching Cubes \cite{lorensen1998marching} on the point cloud represented by depth-projected densified Gaussian primitives. It is noteworthy that, unlike other methods guided by MVS \cite{chen2024mvsplat,cheng2024gaussianpro} or monocular depth estimation-based GS \cite{zhu2023FSGS,chung2023depth,turkulainen2024dnsplatter} training strategies, our approach does not incorporate any external aids. We adopt an approach akin to self-supervision \cite{poggi2020uncertainty,Zhang_2023_CVPR,xiong2023clmvs, chen2023unsupervised}, directly using ground truth (GT) photographs to guide and supervise the back-propagation of the rendered depth \cite{huang20242dgs}. Furthermore, we employ a dynamic densification strategy, which guides the rendering process from coarse to fine detail, effectively enhancing the accuracy and detail of the reconstructed scenes without the need for external depth cues. 
\section{Preliminaries} \small





\subsection{Multi-View Geometric Consistency Filter} \label{geo_consis}

The Geometric Consistency Filter \cite{yao2018mvsnet} is a method for filtering depth maps to remove outliers in background and occluded areas before converting to dense point clouds. This robust filtering strategy is based on photometric and geometric consistencies. Specifically, a reference pixel \(p_1\) with depth \(d_1\) is projected to pixel \(p_i\) in another view, and then reprojected back to the reference image by \(p_i\)’s depth estimation \(d_i\).

\textbf{Photometric consistency} is assessed using a probability map, considering pixels with probabilities below 0.8 as outliers. If the reprojected coordinate \(p_{\text{reproj}}\) satisfy \(|p_{\text{reproj}} - p_1| < 1\), the \(p_1\) is considered photometric consistent.

\textbf{Geometric consistency} involves projecting and reprojecting pixels between views. 
If the reprojected depth \(d_{\text{reproj}}\) satisfy  \(|d_{\text{reproj}} - d_1| / d_1 < 0.01\), the depth estimation \(d_1\) is considered geometric consistent. 

Combine these two-step filtering strategy, requiring depths to be consistent across at least three views, effectively eliminates various outliers.

\subsection{2D Gaussian Splatting}

2DGS \cite{huang20242dgs} is introduced to address the limitations of 3DGS \cite{3dgs} in accurately representing thin surfaces due to multi-view inconsistencies. 2DGS collapses the 3D volume into a set of 2D oriented planar Gaussian disks \cite{pfister2000surfels}, which provide view-consistent geometry. To enhance the quality of reconstructions, two regularization terms are incorporated: depth distortion and normal consistency. The depth distortion term concentrates 2D primitives within a tight range along the ray, while the normal consistency term aligns the rendered normal map with the gradient of the rendered depth. This approach ensures accurate surface representation, achieving state-of-the-art results in geometry reconstruction and novel view synthesis.
\label{sec:methods}

\begin{figure}[!h]
  \centering
   \includegraphics[width=1\linewidth]{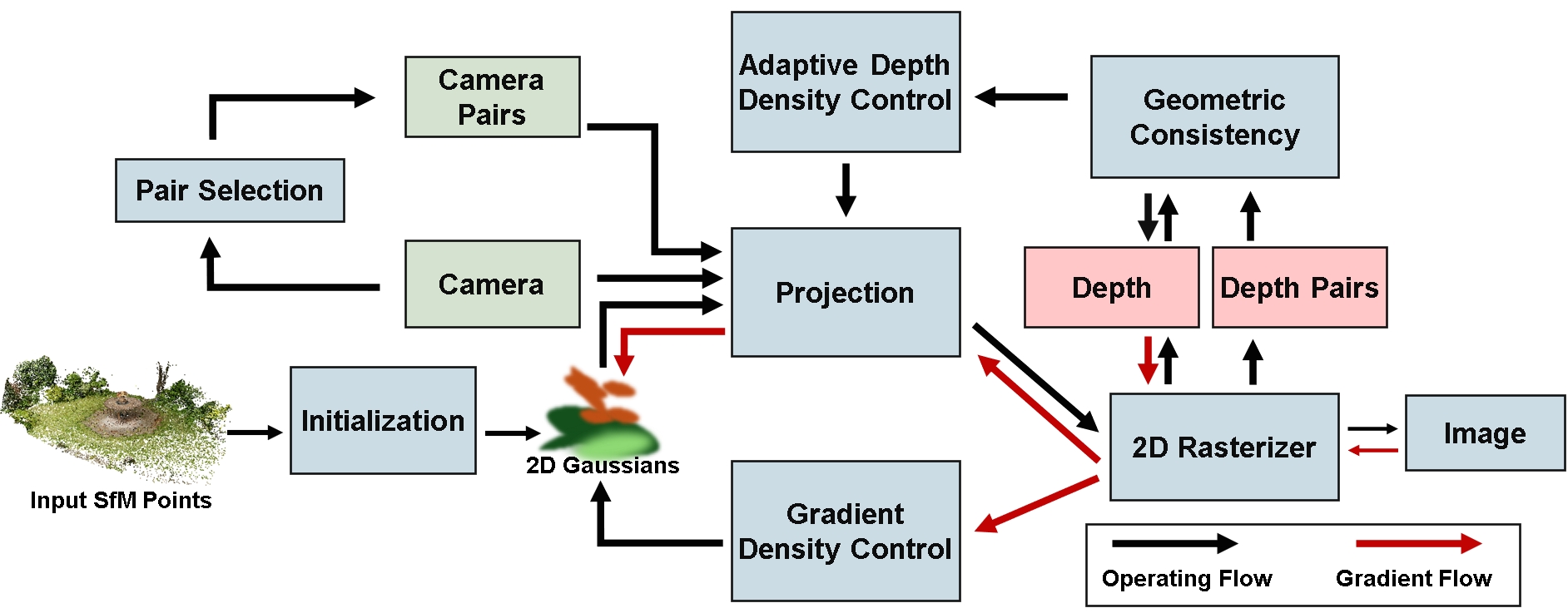}
   \caption{\textbf{Overview of \modelname.} We propose an adaptive densification method based on multi-view geometric consistency, guiding the optimized depth maps to achieve densification through scene projection. Unlike the original GS-based training pipeline, our method initially generates matched photographs based on multi-view principles and guides the optimization and projection of rendered depths during training.
}
   \label{fig:pipeline}
\end{figure}

\section{Methods} 
In this section, we describe the specific process of MVS-Splatting. As illustrated in Figure \ref{fig:pipeline}, we begin by extracting matched photographs from the results of Structure from Motion (SfM). Subsequently, we perform differentiable rendering to obtain rendered images and their corresponding depth. During this process, the rendered images are optimized by recalculating normals in conjunction with the GT images (Section \ref{dnr}). These images and depth data are then subjected to a geometric consistency check, which guides further adaptive depth density control (Section \ref{quantile_depth}). 

\subsection{Depth and Normal Refinement } \label{dnr}

Previous studies assumed that employing rendered depth maps as the ground truth for SDF surface reconstruction \cite{huang20242dgs,zhang2024rade,chen2024pgsr}, which relies significantly on the quality and accuracy of the photometric supervision used during optimization. In conjunction with the results shown in Figure \ref{fig:2dgs_depth_normal}, we propose a method that utilizes normal modification of depth maps. This approach allows for a more rigorous guidance of subsequent multi-view densification processes in Section \ref{quantile_depth}.

\subsubsection{Surface Normal Refinement via Joint or Cross Bilateral Filter}

To enhance the accuracy of the surface normals, we propose an approach that processes the rendered depth maps using a \textit{joint or cross bilateral filter} \cite{yang2013hardware}  and employs a more robust normal calculation method. Initially, \textit{mirror padding} is utilized to ensure that edge pixels to maintain a continuous neighborhood, thereby minimizing edge effects on normal calculation. Following this, the \textit{joint or cross bilateral filter} is employed to smooth the depth map while preserving critical edge details:

\begin{equation}
\resizebox{0.9\hsize}{!}{$
D_{\text{filtered}}(x) = \frac{1}{W(x)} \sum_{x_i } G_{\sigma_s}(\|x_i - x\|) G_{\sigma_r}(\|I(x_i) - I(x)\|) D(x_i)
$} \label{bilateral_filter}
\end{equation}
where \( G_{\sigma_s} \) and \( G_{\sigma_r} \) represent the $3\times3$ spatial and intensity Gaussian kernels, respectively, \( W(x) \) is the normalization factor, \( I(x) \) denotes the pixel value at position \( x \) in the GT image, and \( D(x) \) represents the pixel value at position \( x \) in the rendered depth map.

To compute the normal vectors, we consider the differences between the central pixel and its eight neighboring pixels \cite{yang2018unsupervised}. The depth map is first converted to a 3D point cloud as follows:

\begin{equation}
\resizebox{0.6\hsize}{!}{$
P(u, v) = D(u, v) \cdot K_p^{-1} \cdot [u, v, 1]^T
$} \label{equ:padded_depth_to_3D_Points}
\end{equation}
where $u$ and $v$ are the pixel coordinates in the image plane, and $K_p$ is the intrinsic camera matrix.Subsequently, the normal vectors are computed using the cross products of the differences between 8 neighbor points:

\begin{equation}
\mathbf{n}_i = \frac{\sum_{j=1}^{k=8} (\mathbf{p}_j - \mathbf{p}_i) \times (\mathbf{p}_{j+1} - \mathbf{p}_i)}{\left\| \sum_{j=1}^{k} (\mathbf{p}_j - \mathbf{p}_i) \times (\mathbf{p}_{j+1} - \mathbf{p}_i) \right\|} \label{refined_normal}
\end{equation}
Through these processing steps, our method improves the detail of surface normal estimation, effectively reducing errors introduced by noise and improper boundary handling in the depth maps.

\subsubsection{Depth Maps Refinement via Normals and Image Gradients}
We apply self-supervision ideas for refining depth maps by leveraging normals and image gradients to enhance the accuracy of depth estimation \cite{poggi2020uncertainty,Zhang_2023_CVPR,xiong2023clmvs, chen2023unsupervised,cheng2024gam}. Initially, we apply \textit{mirror padding} to the boundaries of the normal map \(\mathbf{N}\) to provide a more accurate representation of edge details during computational process. 



Next, we compute the depth adjustment factor for each pixel using the normal vector \(\mathbf{n} = (n_x, n_y, n_z)\). The depth adjustment factor \(\eta\) quantifies the influence of the normal vector on the depth value and is formulated as:

\begin{equation}
\eta = \left(\frac{x - c_x}{f_x^{\text{p}}} n_x + \frac{y - c_y}{f_y^{\text{p}}} n_y + n_z \right)
\end{equation}
where \(c_x\) and \(c_y\) denote the coordinates of the camera principal point, and ${f_x^{\text{p}}}$ and ${f_y^{\text{p}}}$ are the updated focal lengths for the padded image. For each pixel \((i, j)\), the corresponding adjustment factor \(\gamma_{i,j}\) is calculated as follows:
\begin{equation}
\gamma_{i,j} = \left(\frac{x_i - c_x}{f_x^{\text{p}}} n_x + \frac{y_j - c_y}{f_y^{\text{p}}} n_y + n_z \right)
\end{equation}

We then update the depth map using the computed adjustment factors:
\begin{equation}
\mathbf{D}_{i,j} = \frac{\eta}{\gamma_{i,j}} \cdot \mathbf{D}_{\text{padded}}
\end{equation}
where \(\mathbf{D}_{i,j}\) represents the updated depth map value at pixel \((i, j)\), and \(\mathbf{D}_{\text{padded}}\) is the padded depth map.
Subsequently, we compute the image gradient:
\begin{equation}
\Delta_{i,j} = \mathbf{I}(x_{i}, y_{j}) - \mathbf{I}(x_{i}, y_{j+1})
\end{equation}
where \(\mathbf{I}(x_i, y_j)\) is the pixel value of the target image at position \((x_i, y_j)\), and \(\Delta_{i,j}\) denotes the gradient in the \((i, j)\) direction. The weights are then derived based on the computed image gradient:

\begin{equation}
w_{i,j} = \exp\left(-\alpha \cdot \left|\Delta_{i,j}\right|\right)
\end{equation}
where \(\alpha\) is the weight decay parameter, and \(w_{i,j}\) signifies the weight in the \((i, j)\) direction. Notably, for each depth value, we compute eight neighboring weights, and the weighted average depth map is obtained as follows:

\begin{equation}
\mathbf{D}_{\text{avg}} = \frac{1}{8}\sum_{i,j}{w_{i,j} \cdot \mathbf{D}_{i,j}} \label{davg}
\end{equation}
where \(\mathbf{D}_{\text{avg}}\) represents the weighted average depth map. Finally, we crop the padded boundaries to obtain the refined depth map.

\subsection{Adaptive Quantile-Based Geometric Consistency Densification} \label{quantile_depth}

In multi-view stereo (MVS) methods \cite{yao2018mvsnet,bleyer2011patchmatch,zhang2023geomvsnet,chen2019pointmvs}, depth maps from different viewpoints are projected into point clouds to achieve 3D reconstruction. Typically, the projected point clouds are concentrated in areas with high confidence, which are mostly located at the center of the sensor or where the depth values are densely populated. A logical way to improve the rendering details is to project the rendered depth map into a point cloud during training and densify the GS scene after initialization \cite{cheng2024gaussianpro}. However, GS-based methods also achieve the most accurate reconstruction in these regions \cite{huang20242dgs,chen2024pgsr,zhang2024rade,Yu2024GOF}. Therefore, the method of only relying on the traditional MVS way to project the depth map will lead to the over-reconstruction of these regions, and at the same time, the regions that need densification will not be optimized effectively.


To address these issues, we propose an adaptive method based on Kernel Density Estimation (KDE) \cite{kim2012robust} and Fast Fourier Transform (FFT) \cite{lee2018single}. This method dynamically estimates the depth map distribution and determines the quantile thresholds. The depth map is adaptively divided into \textit{near}, \textit{mid}, and \textit{far} regions, with densification focused on \textit{near-camera} and \textit{far-camera} areas. By dynamically adjusting the segmentation of each depth map, we determine the corresponding densification strategy.

\subsubsection{Adaptive Quantile Calculation}\label{aqc}
The objective of KDE is to obtain a smoothed estimate of the probability density distribution of the given depth data $D$ using a Gaussian kernel function. To compute the KDE efficiently, we use the Fast Fourier Transform (FFT) to speed up the convolution operation. First, we perform FFT on the kernel function to obtain $K(f)$ and depth map to obtain $D(f)$. Next, we multiply $K(f)$ and $D(f)$ to obtain the frequency domain representation of the convolution result $(K \ast D)(f)$. Then, we perform inverse FFT on the frequency domain result to obtain the convolution result $(K \ast D)(t)$. This can be represented mathematically as:
\begin{equation}
(K \ast D)(t) = \mathcal{F}^{-1} \left( \mathcal{F}(K) \cdot \mathcal{F}(D) \right)
\end{equation}
where $F(\cdot)$ represents the FFT oerator. After this, we obtain the smoothed depth distribution, allowing us to compute the cumulative distribution function (CDF) and determine the required quantiles:
\begin{equation}
\text{CDF}(x) = \int_{-\infty}^{x} \hat{f}(t) \, dt
\end{equation}
where $\hat{f}(t)$ is the density function obtained from the KDE calculation. We further derive the specific quantiles for segmenting the depth map into near, mid, and far regions. We calculate the desired quantiles from the KDE result. The quantile calculation is given by:

\begin{equation}
q = \text{CDF}^{-1}(p)
\end{equation}
\begin{figure}[t]
  \centering
   \includegraphics[width=\linewidth]{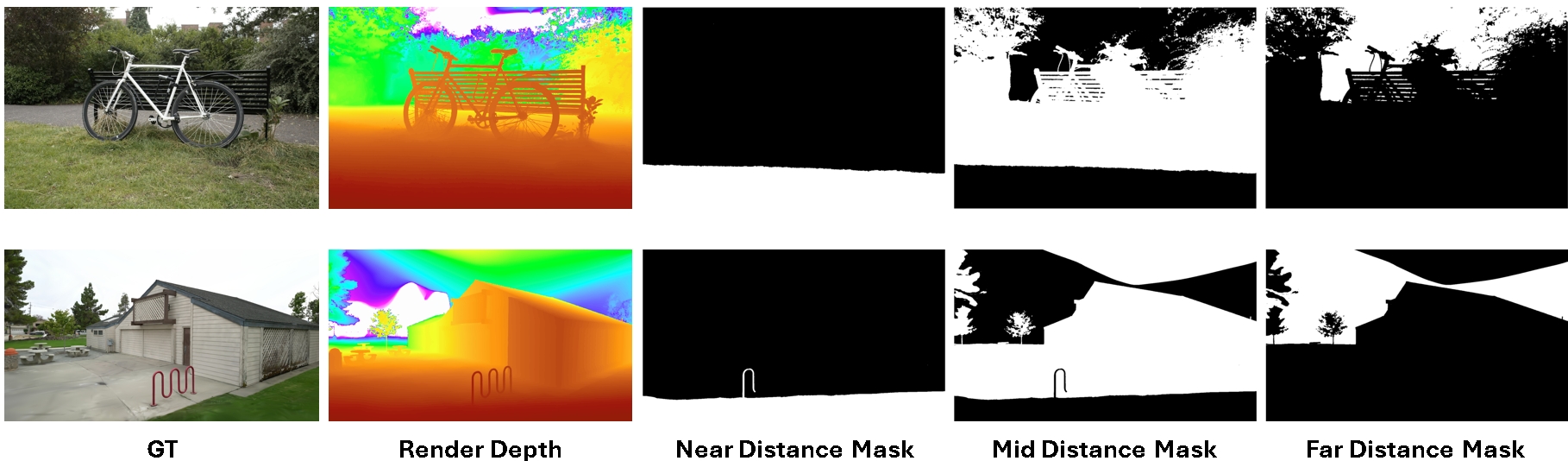}
   \caption{\textbf{Adaptive Quantile-Based Segmentation.} We propose an adaptive method based on Kernel Density Estimation (KDE) and Fast Fourier Transform (FFT). This method dynamically estimates the depth map distribution and determines the quantile threshold for the depth based densification in the next step. From top to bottom: Mip-NeRF 360 Bicycle scene and Tanks and Temple Bran scene.}
   \label{fig:q_mask}
\end{figure}
where $p$ is the rangs of three different quantile probability (near, mid, far). Figure \ref{fig:q_mask} shows the example depth division results for two different outdoor scenes. Specifically, we dynamically adjust the quantile probability $P_{range}$ of every render depth in each iteration by defining the quantile probability range $p$ to determine the adaptive quantile threshold. In this way, we are able to flexibly adjust the division of the three regions near, middle and far according to the specifics of each depth map, thus achieving more effective densification.

\subsubsection{Densification Strategy Based on Multi-View Geometric Consistency} \label{dd}
\begin{figure}[!h]
  \centering
   \includegraphics[width=\linewidth]{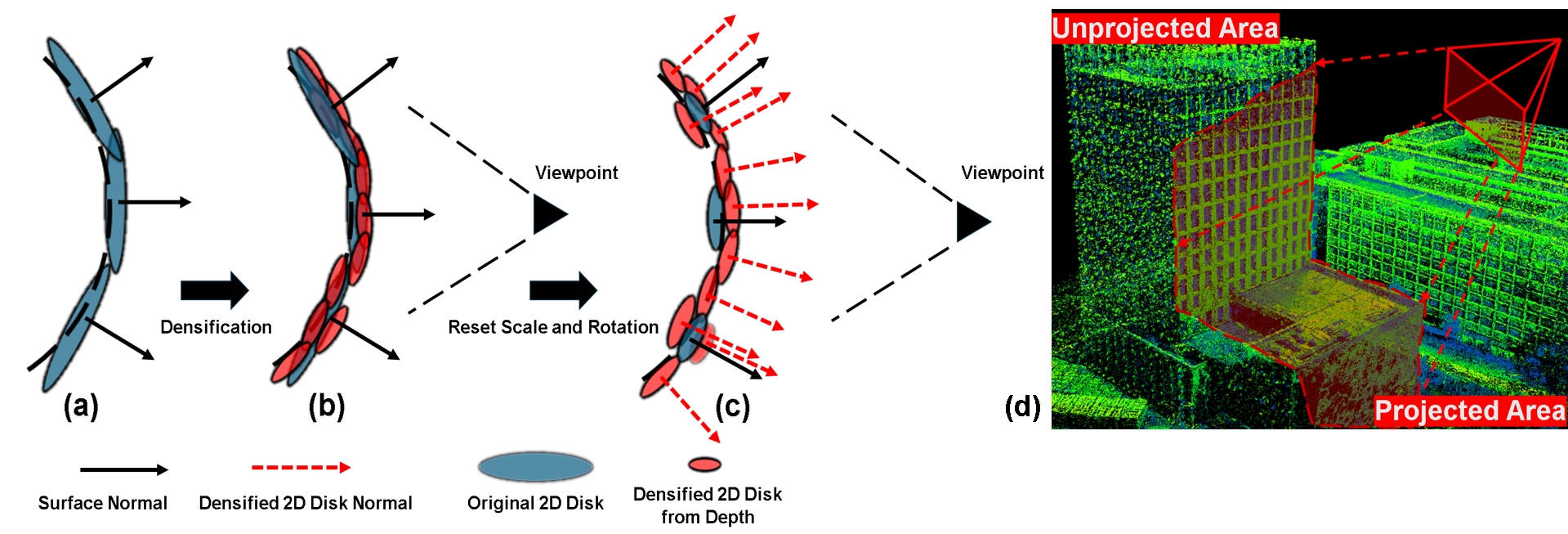}
   \caption{\small\textbf{Densification Strategy.} We present a depth map-based densification method to enhance rendering and mesh extraction quality. (a) We first identify under-reconstructed areas and acquire surface normals. (b) Through multi-view geometric consistency, depth maps are projected onto these areas. (c) Using surface normals, the orientation of projected primitives is adjusted to be perpendicular to the normals, enhancing alignment. The scale of all primitives is reinitialized for improved rendering accuracy. (d) An example from a single viewpoint shows that the projected area achieves more uniform and precise densification compared to unprojected areas.}
   \label{fig:densify_pipeline}
\end{figure}

Unlike previous work \cite{cheng2024gaussianpro}, our depth map-based densification method will not mainly focus on regions with high geometric consistency. We follow the densification approach of 3DGS \cite{3dgs}, that is, focusing on regions with less geometric features. Specifically, as shown in Figure \ref{fig:densify_pipeline}, this is an adaptive projection method based on geometric and photometric consistency in Section \ref{geo_consis}, which combines the advantages of multi-view depth maps and 2DGS \cite{huang20242dgs} surface fitting, and improves the overall quality and detail reconstruction of the point cloud by adjusting the geometric consistency tolerance for different regions.

\textbf{Densification Strategy}: 
To avoid over-reconstruction, we first determine whether to perform densification based on the relative scale of each 2D Gaussian primitive in the scene. Specifically, we set a threshold, and densification is triggered only when the size of the primitive exceeds this threshold. 
During the densification process, to determine the number of projection points, we follow the calculations for depth consistency and photometric consistency in Section \ref{geo_consis}. In the previous Section \ref{quantile_depth}, we divide the scene into three regions: \textit{near}, \textit{middle}, and \textit{far}. We then optimize the geometric consistency tolerance settings to enhance reconstruction performance. For the \textit{near} and \textit{far} regions, which are typically under-reconstructed due to limitations of sensor physical properties and camera-to-object distance respectively, the main densification will occur in these two regions. By increasing the geometric consistency tolerance, more points can pass the consistency check, thereby densifying the under-reconstructed areas. In the \textit{middle} region, which is usually well-reconstructed through Gaussian surface fitting, we reduce the geometric consistency tolerance. This approach aims to supplement existing reconstruction details without excessively modifying already densely reconstructed areas, thus avoiding the introduction of new errors. Note that our geometric consistency check and projection are based on the refined depth processed through normal and image gradient, as indicated in Equation \ref{davg}.

\textbf{Initialization of Projection Primitives:}
Unlike depth densification in 3DGS-based studies \cite{cheng2024gaussianpro, chen2024mvsplat}, where randomly initializing the rotation  does not significantly impact the final rendering and reconstruction results. As illustrated in 2DGS, the unique property of 2D Gaussian primitives requires to address the degeneracy problem \cite{huang20242dgs}, that is, the need for a reasonable design to initialize the rotation of the depth projected primitives. In our approach, the rotation for each point is initialized to align with the normal vectors of the surfaces they represent. 
For each point we estimate the normal vector $n_i$ using Equation \ref{refined_normal}. Next, we represent the rotation using the axis-angle format. The axis of rotation is given by the cross product of the normal vector $n_i$ and a reference vector $r$, and the angle of rotation is the arccosine of their dot product:
\begin{equation}
    \theta_i=cos^{-1}(n_i \cdot r)
\end{equation}
We then convert the axis-angle representation to a rotation matrix $R_i$ This is done using the exponential map from the axis-angle representation to the rotation matrix:
\begin{equation}
    R_i = exp([\theta _i]_\times )
\end{equation}
where $[\theta _i]_\times$ is the skew-symmetric matrix of $\theta _i$.

\begin{figure}[!h]
  \centering
   \includegraphics[width=\linewidth]{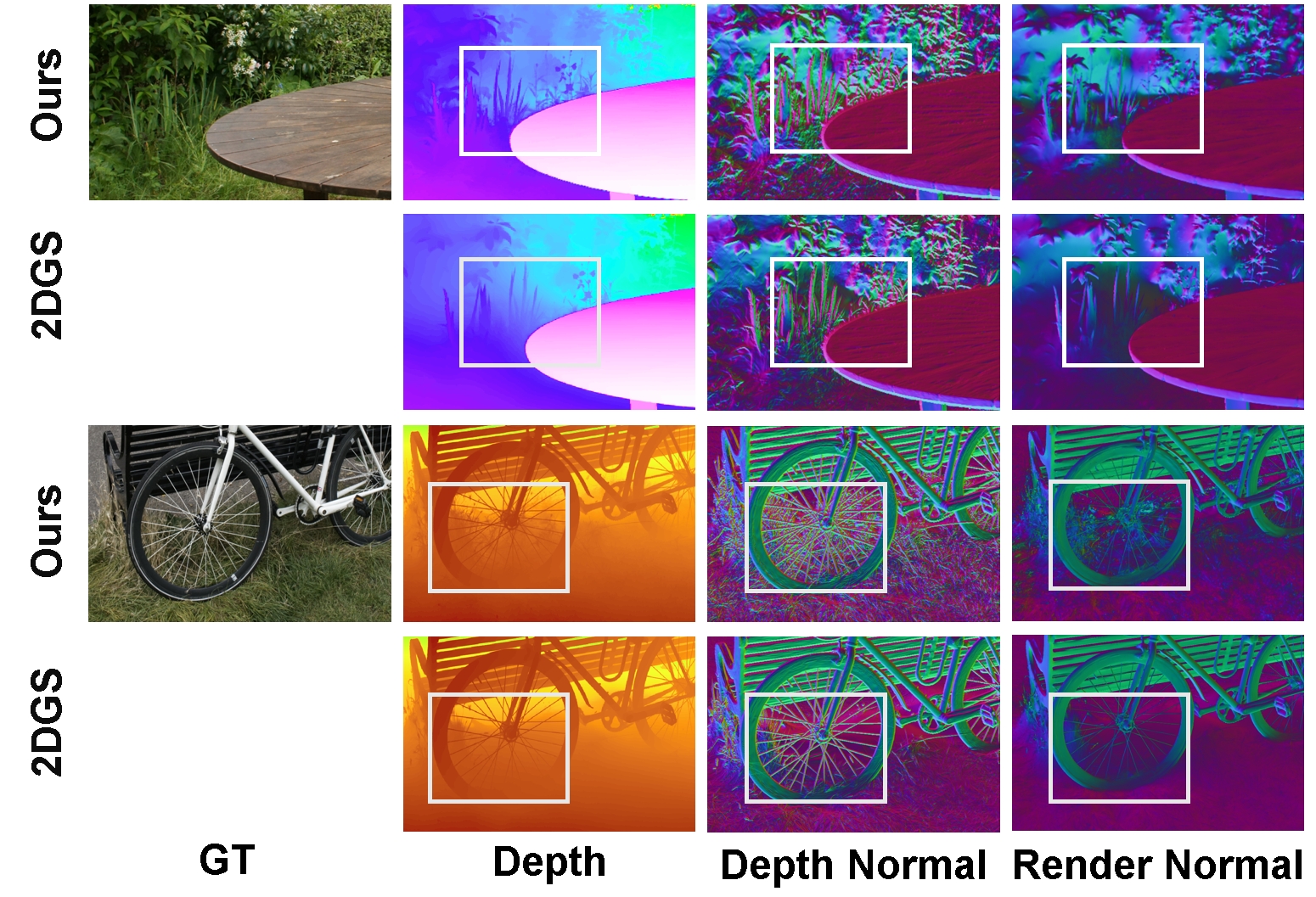}
   \caption{\textbf{Depth and Normals comparsion.}  It is evident from these comparisons that our final results exhibit more comprehensive detail representation compared to 2DGS.}
   \label{fig:depth_normal_compare}
\end{figure}

Only initialize the scale of the projected points will lead the original maximum scale of the 2D Gaussian primitives remain unchanged. The areas requiring densification will be covered or influenced by the original largest 2D Gaussian primitives, preventing the learning of appropriate features and opacity. This will lead to persistent depth calculation errors during training. Therefore, it is necessary to locally reinitialize the scale in the densification areas. Specifically, after every depth densification iteration, we reinitialize the scale of each 2D Gaussian primitive in the selected areas based on the distances between neighboring points. This ensures that the scale of the original 2D Gaussian primitives in the areas requiring reconstruction is also adjusted, accurately representing the details of different areas. Figure \ref {fig:depth_normal_compare} showcases a visualization example after densification, illustrating that our method is capable of rendering more detailed depth maps and two types of normal maps.

\subsection{Joint Loss Function}
Our objective is to effectively constrain the rendered depth maps using GT images without introducing additional methods such as MVS or monocular depth estimators. Inspired by several unsupervised MVS works and depth-supervised GS methods \cite{poggi2020uncertainty,Zhang_2023_CVPR,xiong2023clmvs, chen2023unsupervised,cheng2024gam}, we design a joint loss function to provide effective supervision for the depth maps and render images.

\textbf{Edge Aware Depth Loss:}
To ensure edge consistency between the GT image and the depth map, we introduce an edge loss function \cite{turkulainen2024dnsplatter}. We use the Scharr operator \cite{Scharr2000OptimalOI} to compute the gradients of the image and depth map in the x and y directions, respectively, and combine these gradients to form an edge image. The equation for the edge aware loss that incorporates the edge detection component is as follows:
\begin{equation}\resizebox{0.85\hsize}{!}{$
    \mathcal{L}_{edge} =\lambda_x\cdot log(1+\left | D-D_{evl} \right | )_x+\lambda_y\cdot log(1+\left | D-D_{avg} \right | )_y
    $}
\end{equation}
where $D$ is the render depth map, $D_{avg}$ is the refined depth map from Equation \ref{davg}, and $\lambda_x = exp( -\left | E_{rgb}(x)-E_{rgb}(x+1) \right |)$ and $\lambda_y = exp( -\left | E_{rgb}(y)-E_{rgb}(y+1) \right |)$ are weights computed from the edges of the RGB image, which guide the loss function to be more sensitive to edges in the image. $E_{rgb}$ denotes the edge map of the RGB image obtained using Scharr edge detection \cite{Scharr2000OptimalOI} and non-maximum suppression \cite{8100168}.

\textbf{Feature Aware Depth Loss:}
Although depth maps do not contain color information, the texture features (e.g., surface details) in the GT image and the depth map should be similar \cite{chen2023unsupervised,poggi2020uncertainty}. Therefore, we leverage a pre-trained VGG network features and computing the discrepancy between them:
\begin{equation}
    \mathcal{L}_{feat} = \left \| \phi (I_{GT})- \phi (D) \right \| 
\end{equation}
Where $\phi$ represents the feature extraction function of the pre-trained VGG network.

\textbf{Final Loss:} Finally, the total loss function is as follows:

\begin{equation}
    \mathcal{L} = \mathcal{L}_{rgb}+\alpha \mathcal{L}_{edge}+\beta \mathcal{L}_{feat}+\gamma \mathcal{L}_{n}
\end{equation}
where $\mathcal{L}_{rgb}$ is the original loss function from 3DGS \cite{3dgs}, and $\mathcal{L}_{n}$ is the normal consistenct loss from 2dgs \cite{huang20242dgs}.

\section{Experiments}
\label{sec:exp}
\small
In this section, we  conduct a comprehensive analysis of MVG-Splatting, including evaluations of both novel view synthesis (NVS) rendering quality and mesh extraction quality. To verify the efficacy of each component within our approach, detailed ablation studies will be performed for each module. 
\subsection{Implementation}

\begin{figure*}[!h]
  \centering
   \includegraphics[width=\linewidth]{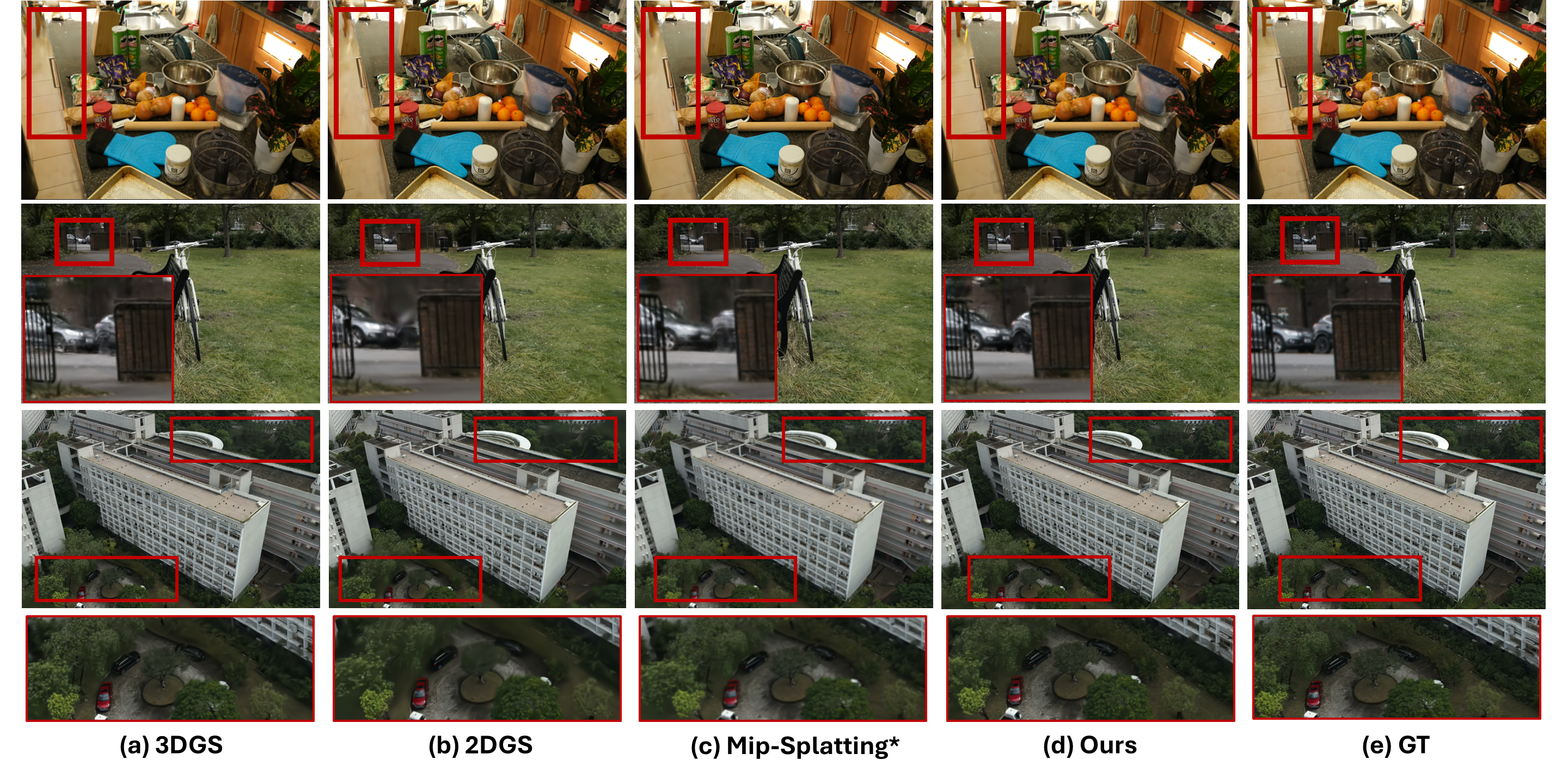}
   \caption{\textbf{Novel View Synthesis Comparison.} Comparison of NVS results on the Mip NeRF 360 ]\cite{barron2022mipnerf360} and Urbanscene 3D \cite{UrbanScene3D} datasets. From top to bottom: \textit{counter}, \textit{bicycle}, and \textit{Artsci}. We selected \textit{near} and \textit{far} areas for comparison, demonstrating that our densification method effectively densifies under-reconstructed areas, thereby enhancing the rendering quality.}
   \label{fig:nvs}
\end{figure*}
Building upon the foundation of 2DGS \cite{huang20242dgs}, we develop the MVG-Splatting, which incorporates the depth-normal optimization based on image gradients, adaptive depth projection densification, and depth regularization as previously outlined. Initially, each test scene undergoes 15,000 iterations of the original adaptive densification proposed by 3DGS \cite{3dgs}, followed by 5,000 iterations of optimization. This procedure aims to generate detailed initial renderings of depth and surface normals for the scenes. Subsequently, leveraging the optimized normals and depth maps, the scenes underwent 10,000 iterations of depth projection densification, with densification intervals set at every 100 iterations. For the projection areas segmented using an Adaptive Quantile-Based calculation, we uniformly set the photometric and geometric consistency \ref{geo_consis} thresholds at $p_{reproj}$=1 and $d_{reproj}$=0.01 for the \textit{near} and \textit{far} areas, respectively, and at $p_{reproj}$=1 and $d_{reproj}$=0.001 for the \textit{mid} areas. Upon completion of the training, we employ a custom Marching Cubes \cite{lorensen1998marching} algorithm to extract surfaces from the densely generated Gaussian point clouds, with a base voxel size set at 0.003. Based on local point cloud density, the voxel size dynamically adjusts between 0.005 and 0.001. All experiments were conducted on an NVIDIA RTX 4090 GPU.

\subsection{Novel View Synthesis}
\begin{table}[h]
    \centering
    \scriptsize
    \begin{tabular}{l|ccc}
        
        & \textbf{PSNR} $\uparrow$ & \textbf{SSIM} $\uparrow$ & \textbf{LPIPS} $\downarrow$ \\
        \hline
        NeRF \cite{mildenhall2021nerf}  & 23.85 & 0.605 & 0.451 \\
        Mip-NeRF \cite{barron2021mipnerf}  & 24.04 & 0.616 & 0.441 \\
        NeRF++ \cite{zhang2020nerf++}  & 25.11 & 0.676 & 0.375 \\
        Plenoxels \cite{fridovich2022plenoxels}  & 23.08 & 0.628 & 0.463 \\
        Instant NGP \cite{mueller2022instantngp}  & 25.68 & 0.705 & 0.302 \\
        Mip-NeRF 360 \cite{barron2022mipnerf360}  & 27.57 & 0.789 & 0.234 \\
        \hline
        3DGS \cite{3dgs}  & 27.21 & 0.825 & 0.214 \\
        2DGS \cite{huang20242dgs}  & 27.17 & 0.816 & 0.210 \\
        Mip-Splatting \cite{yu2024mipsplatting}& \cellcolor[HTML]{FFFFC7}27.79 & \cellcolor[HTML]{FFFFC7}0.827 & \cellcolor[HTML]{FFFFC7}0.203 \\
        Mip-Splatting* \cite{yu2024mipsplatting} & \cellcolor[HTML]{FD6864}27.88 & \cellcolor[HTML]{FD6864}0.835 & \cellcolor[HTML]{FFCE93} 0.198 \\
        MVG-Splatting (Ours)  & \cellcolor[HTML]{FFCE93}27.84 & \cellcolor[HTML]{FFCE93}0.834 & \cellcolor[HTML]{FD6864}0.196 \\
        \hline
    \end{tabular}
    \caption{Novel view synthesis comparison of different models on \textbf{Mip-NeRF 360} \cite{barron2022mipnerf360} dataset with PSNR, SSIM, and LPIPS metrics. Mip-Splatting* means we report Mip-Splatting \cite{yu2024mipsplatting} metrics with GOF \cite{Yu2024GOF} densification.}
    \label{table:mip_nerf_360_nvs}
\end{table}

\begin{table*}[!t]
    \centering
    \scriptsize
    \begin{tabular}{l|ccccccccccccccc|c|c}
        \hline
        \textbf{Method} & 24 & 34 & 40 & 55 & 63 & 65 & 69 & 83 & 97 & 105 & 106 & 110 & 114 & 118 & 122 & \textbf{Mean} & \textbf{Time} \\
        \hline
        \textbf{NeRF} \cite{mildenhall2021nerf}  &  1.90& 1.60& 1.85& 0.58& 2.28& 1.27& 1.47& 1.67& 2.05& 1.07& 0.88& 2.53& 1.06& 1.15& 0.96& 1.49 & $>$12.0h \\
        
        \textbf{VolSDF} \cite{yariv2021volume} & 1.14& 1.26 &0.81&0.49& 1.25&\cellcolor{orange!30} 0.70&\cellcolor{orange!30} 0.72& 1.29& \cellcolor{orange!30}1.18&\cellcolor{red!30} 0.70& \cellcolor{orange!30}0.66& \cellcolor{red!30}1.08 &\cellcolor{yellow!30}0.42&\cellcolor{orange!30} 0.61& 0.55& 0.86& $>$12.0h \\
        
        \textbf{NeuS} \cite{wang2021neus}  & 1.00 & 1.37 &0.93& 0.43&\cellcolor{yellow!30} 1.10 &\cellcolor{red!30}0.65& \cellcolor{red!30}0.57& 1.48& \cellcolor{red!30}1.09& 0.83& \cellcolor{red!30}0.52&\cellcolor{orange!30} 1.20&\cellcolor{red!30} 0.35&\cellcolor{red!30} 0.49& 0.54& 0.84& $>$12.0h\\
        \hline
        \textbf{SuGaR} \cite{guedon2023sugar}  &  1.47 & 1.33& 1.13 &0.61& 2.25& 1.71& 1.15& 1.63 &1.62 &1.07& 0.79 &2.45 &0.98& 0.88& 0.79& 1.33& 1h \\
        \textbf{3DGS} \cite{3dgs} & 2.14& 1.53& 2.08& 1.68& 3.49& 2.21 &1.43 &2.07 &2.22 &1.75& 1.79& 2.55 &1.53& 1.52& 1.50& 1.96& \cellcolor{red!30}11.2m \\
        \textbf{2DGS} \cite{huang20242dgs} & \cellcolor{yellow!30} 0.48 & \cellcolor{yellow!30} 0.91 &\cellcolor{yellow!30}0.39&\cellcolor{orange!30} 0.39&\cellcolor{orange!30} 1.01& 0.83& 0.81 &\cellcolor{yellow!30}1.36& \cellcolor{yellow!30}1.27& 0.76&\cellcolor{yellow!30} 0.70& 1.40&\cellcolor{orange!30} 0.40&0.76& \cellcolor{yellow!30}0.52& \cellcolor{yellow!30}0.80& \cellcolor{orange!30}18.8m \\
        
        \textbf{GOF} \cite{Yu2024GOF} & \cellcolor{orange!30} 0.50 & \cellcolor{red!30}0.82 &\cellcolor{orange!30}0.37&\cellcolor{red!30} 0.37& 1.12& \cellcolor{yellow!30}0.74&\cellcolor{yellow!30} 0.73 &\cellcolor{red!30}1.18& 1.29& \cellcolor{orange!30}0.68& 0.77& 1.90& 0.42&\cellcolor{yellow!30} 0.66&\cellcolor{orange!30} 0.49& \cellcolor{red!30}0.74& 50m \\
        
        \textbf{MvG-Splatting} (Ours) & \cellcolor{red!30}0.46 & \cellcolor{orange!30}0.84 & \cellcolor{red!30}0.35 & \cellcolor{red!30}0.37 & \cellcolor{red!30}0.99 & 0.78 & \cellcolor{yellow!30}0.73 & \cellcolor{orange!30} 1.19 & \cellcolor{yellow!30}1.27 & \cellcolor{yellow!30}0.74 & 0.72 & \cellcolor{yellow!30}1.27 & \cellcolor{yellow!30}0.40 & \cellcolor{yellow!30} 0.66 & \cellcolor{red!30}0.47 &\cellcolor{orange!30} 0.75 &  \cellcolor{yellow!30}25m\\
        \hline
    \end{tabular}
    \caption{Quantitative comparison on the DTU Dataset \cite{jensen2014dtu}. We report chamfer distance and training time.}
    \label{DTU}
\end{table*}
To evaluate MVG-Splatting, we conducted comparative analyses on the Mip-NeRF 360 dataset \cite{barron2022mipnerf360} against advanced methods based on NeRF and GS. Quantitative evaluation results are displayed in Table \ref{table:mip_nerf_360_nvs}. Compared to 2DGS \cite{huang20242dgs}, our approach significantly enhances rendering outcomes and shows a slight lead over SOTA GS-based rendering method Mip-Splatting \cite{yu2024mipsplatting}. Notably, in the LPIPS assessment, our method outperforms all others, including Mip-splatting* \cite{yu2024mipsplatting} that utilizing GOF \cite{Yu2024GOF} densification strategies. We attribute these enhancements primarily to our quantile-based densification strategy.
Figure \ref{fig:nvs} presents three sets of \textit{indoor}, \textit{outdoor }and \textit{aerial} examples from the test set. It is evident that our method is better in the rendering of the segmented \textit{near} and \textit{far} areas compared to other methods. This advantage stems from our relatively lenient projection strategies in these regions, which still effectively initialize Gaussian primitives at accurate positions. This demonstrates that our method not only preserves high-quality visual fidelity but also excels in precise geometrical localization and consistency across varied viewing conditions.

\subsection{Mesh Extraction}

We conducted a mesh comparison on the DTU dataset \cite{jensen2014dtu}, as presented in Table \ref{DTU}. In comparisons against GS-based methods, our approach outperformed 3DGS \cite{3dgs}, SuGaR \cite{guedon2023sugar}, and 2DGS \cite{huang20242dgs}, and was on par with the SOTA method GOF \cite{Yu2024GOF} in average evaluation metrics. Notably, in the evaluation of training duration, our method slightly exceeded 2DGS \cite{huang20242dgs} but was significantly faster than both SuGaR \cite{guedon2023sugar} and GOF \cite{Yu2024GOF}. Table \ref{tab:tnt} presents the quantitative results on the Tanks and Temples dataset \cite{knapitsch2017tanks}, where our method continued to achieve second place in overall performance, and significantly reduced the training time compared to SuGaR \cite{guedon2023sugar} and GOF \cite{Yu2024GOF}. 

\begin{figure}[!h]
  \centering
   \includegraphics[width=\linewidth]{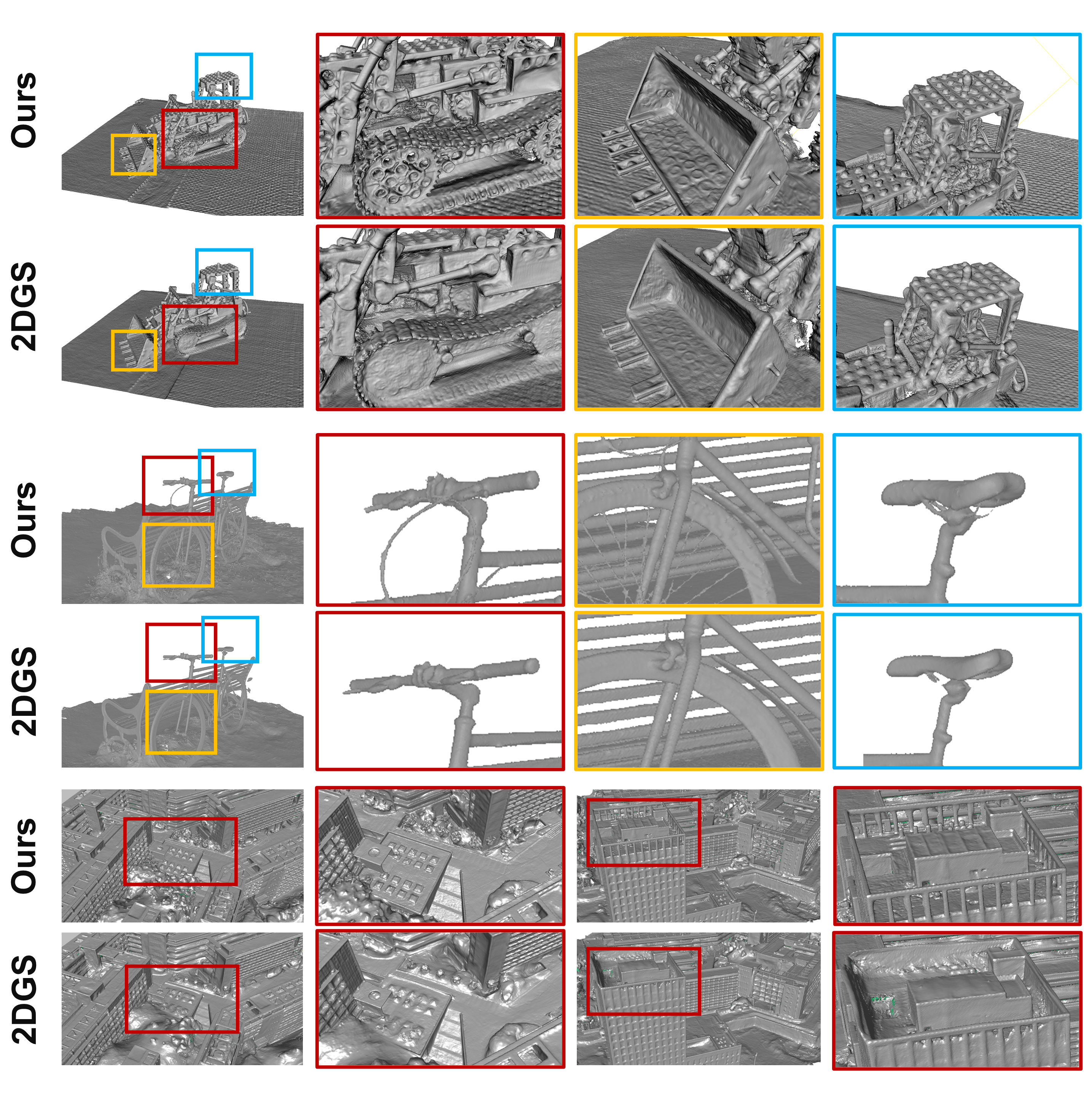}
   \caption{\textbf{In-the-Wild Mesh Compairson.} Comparison of meshes extracted using our method versus those obtained with 2DGS  employing TSDF.  Our method more effectively reconstructs a greater amount of surface details from the images, showcasing superior detail retrieval capabilities. }
   \label{fig:mesh}
\end{figure}

Overall, our approach represents a more refined improvement based on 2DGS \cite{huang20242dgs}. The visiual results in Figure \ref{fig:mesh} confirm that our method can achieve a noticeable enhancement in mesh extraction quality with minimal loss in training efficiency compared to 2DGS \cite{huang20242dgs}. This demonstrates the effectiveness of our refinements in optimizing both the computational aspects and the qualitative outcomes of the mesh reconstruction process. This balance is crucial for in-the-wild applications where both speed and precision are valued.

\begin{table}[h]
    \centering
    \scriptsize
    \begin{tabular} [width=\linewidth] {l|cc|cccc}
        & \textbf{NeuS} & \textbf{Geo-Neus}  & \textbf{SuGaR} & \textbf{2DGS} & \textbf{GOF}&\textbf{Ours} \\
        \hline
        Barn & 0.29 & 0.33  & 0.14 & \cellcolor{yellow!30}0.36 & \cellcolor{red!30}0.51 & \cellcolor{orange!30}0.45 \\
        Caterpillar & \cellcolor{yellow!30}0.29 & 0.26  & 0.16 & 0.23 &\cellcolor{red!30}0.41 &\cellcolor{orange!30}0.37 \\
        Courthouse &\cellcolor{yellow!30}0.17 & 0.12  & 0.08 & 0.13  &\cellcolor{red!30}0.28 & \cellcolor{orange!30}0.27 \\
        Ignatius &\cellcolor{red!30}0.83 &\cellcolor{orange!30} 0.72  & 0.33 & 0.44 &\cellcolor{yellow!30}0.68 & 0.59 \\
        Meetingroom & \cellcolor{orange!30}0.24 & 0.20  & 0.15 & 0.16 & \cellcolor{red!30}0.28 & \cellcolor{yellow!30}0.23 \\
        Truck & \cellcolor{orange!30}0.45 & \cellcolor{orange!30}0.45  & 0.26 & 26 & \cellcolor{red!30}0.59 &  \cellcolor{yellow!30}0.39\\
        \hline
        Mean & \cellcolor{orange!30}0.38 & \cellcolor{yellow!30}0.35  & 0.19 & 0.30 & \cellcolor{red!30}0.46 &  \cellcolor{orange!30}0.38\\
        Time & $>$24h & $>$24h  & \cellcolor{yellow!30}$>$1h & \cellcolor{red!30}34.2 m & \cellcolor{yellow!30}$>$1h & \cellcolor{orange!30}49.9 m \\
    \end{tabular}
    \caption{Comparison of various methods on Tanks and Temples datasets \cite{knapitsch2017tanks}. We report F-1 score and average training time.}
    \label{tab:tnt}
\end{table}

\subsection{Ablations}

In this section, we will conduct ablation studies primarily aimed at validating the effectiveness of the various modules we have proposed, in terms of their impact on NVS and mesh extraction capabilities. The analysis results are presented in Table \ref{table:mip_nerf_360_albation} and \ref{table:mesh_abl}. 

\textbf{Novel View Synthesis:}
The ablation study on Mip-NeRF 360 dataset \cite{barron2022mipnerf360} detailed in the Table \ref{table:mip_nerf_360_albation} rigorously assesses the contributions of critical components within MVG-Splatting. The results from Table \ref{table:mip_nerf_360_albation} A indicate that removing the depth-normal refinement (D-NR) in Section \ref{dnr} leads to a noticeable degradation in rendering quality. Table \ref{table:mip_nerf_360_albation} B, which lacks both depth densification (DD) in Section \ref{dd} and adaptive quantile-based calculation (AQC) in Section \ref{aqc}, shows further reduction in render qiality. In Table \ref{table:mip_nerf_360_albation} C, the removal of AQC alone results in better performance compared to \ref{table:mip_nerf_360_albation} B but still underperforms relative to the full model. This underscores DD's role in enhancing detail through more targeted and efficient densification.

\begin{table}[h]
    \centering
    \setlength{\tabcolsep}{4pt}
    \scriptsize
    \begin{tabular}{l|ccc|cc}
        
        & \textbf{PSNR} $\uparrow$ & \textbf{SSIM} $\uparrow$ & \textbf{LPIPS} $\downarrow$ & \textbf{Time} $\downarrow$ & \textbf{Size (MB)}$\downarrow$ \\
        \hline
        A. w/o D-NR & 27.49 & 0.819 & 0.219 & 45 m & 1279 \\
        B. w/o DD and AQC & 27.19 & 0.815 & 0.239 & 36 m & 899 \\
        C. w/o AQC & 27.69 & 0.820 & 0.205 & 67 m & 1874 \\
        \hline
        D. Ours full   & 27.80 & 0.834 & 0.196 & 54 m & 1282 \\
        \hline
    \end{tabular}
    \caption{\textbf{Novel View Synthesis Ablation Studies.} We conducted ablation experiments on the Mip-NeRF 360 dataset \cite{barron2022mipnerf360}. We reported metrics including PSNR, 
 SSIM, LPIPS, training time, and the storage  size for Gaussian point clouds.  }
    \label{table:mip_nerf_360_albation}
\end{table}

\begin{table}[h]
    \centering
    \setlength{\tabcolsep}{4pt}
    \scriptsize
    \begin{tabular}{l|ccc}
       
        & \textbf{Accuracy $\downarrow$} & \textbf{Completion $\downarrow$} & \textbf{Average $\downarrow$} \\
        \hline
        A. w/o normal consistency & 1.24 & 1.15 & 1.19 \\
        B. w/o depth edge & 0.88 & 0.87 & 0.88 \\
        C. w/o depth feat. & 0.77 & 0.99 & 0.88 \\
        \hline
        D. w/ TSDF & 0.76 & 0.83 & 0.79 \\
        E. Ours full & 0.72 & 0.78 & 0.75 \\
        \hline
    \end{tabular}
    \caption{\textbf{Mesh Extraction Ablation Studies.} We conducted ablation experiments on the Tanks and Temples dataset \cite{knapitsch2017tanks}. We report accuracy, completion, and average metrics.}
    \label{table:mesh_abl}
\end{table}
\textbf{Mesh Extraction:} In the provided Table \ref{table:mesh_abl}, an ablation study evaluates the efficacy of various regularization components within our mesh reconstruction methods. In Table \ref{table:mesh_abl} A, where normal consistency regularization is absent, suggests that normal consistency is vital for maintaining the geometric coherence of the reconstructed surfaces. Table \ref{table:mesh_abl} B shows that depth edge handling improves sharpness and definition at object boundaries, its impact is less critical than normal consistency but still significantly beneficial. Removing depth feature (Table \ref{table:mesh_abl} C) regularization leads to mixed results; accuracy improves, but completion worsens, resulting in an unchanged average score compared to Table \ref{table:mesh_abl} B. Table \ref{table:mesh_abl} D shows using TSDF for surface extraction yield good results, however, it still cannot enhance surface detail when comparing with ours full model (Table \ref{table:mesh_abl} E).
\section{Conclusion}
\small
\label{sec:conclusion}
We propose MVG-Splatting, a method that uses optimised normal calculations in conjunction with image gradients. The proposed method effectively resolves the inconsistencies often found in depth estimations using GS-based methods. This refinement ensures a more accurate and visually consistent rendering. Additionally, our adaptive quantile-based densification strategy dynamically enhances the rendering and mesh quality from coarse to detailed, aligning with the depth variations observed across multiple views. Experimental results validate that our approach not only mitigates the decline in rendering quality caused by depth discrepancies but also leverages dense Gaussian point clouds for mesh extraction via the Marching Cubes algorithm, substantially improving both the fidelity and accuracy of the reconstructed models.

\textbf{Limitations:} Despite its strengths, our method exhibits limitations, particularly in processing extremely large datasets where computational overhead can become substantial. The adaptive quantile-based approach, while effective, requires careful calibration to balance between detail enhancement and computational efficiency.  Moreover, the dependency on high-quality image gradients and precise normal calculations might limit the method's applicability in scenarios with poor image quality (such as overexposure, back-lighting, or blurriness), potentially affecting the robustness and versatility of the 3D reconstruction process. Future work can focus on the above limitations.

{
    \scriptsize
    \bibliographystyle{ieeenat_fullname}
    \bibliography{main}
}

 \clearpage
\setcounter{page}{1}
\maketitlesupplementary
\small
\subsection{DETAILS OF DEPTH-NORMAL REFINEMENT}
Recall that our normal refinement consider the differences
between the central pixel and its eight neighboring pixels of each rendered surface depth map in Equation \ref{refined_normal}. Before this computation, we apply \textit{mirror padding} to better perform edge recovery of the normal maps. The depth map is first converted to a 3D point cloud as repersent in Equation \ref{equ:padded_depth_to_3D_Points} :
\begin{equation}
\resizebox{0.6\hsize}{!}{$
\mathbf{P}(u, v) = D(u, v) \cdot K_p^{-1} \cdot [u, v, 1]^T
$} 
\end{equation}
Note that due to \textit{mirror padding}, the dimensions of the depth map are altered to \(H + 2n\) and \(W + 2n\). Consequently, the intrinsic matrix \(\mathbf{K}_{\text{p}}\) must be updated to account for these changes:

\begin{equation}
\resizebox{0.5\hsize}{!}{$
\mathbf{K}_{\text{p}} = \begin{bmatrix} f_x^{\text{p}} & 0 & (W + 2n)/2 \\ 0 & f_y^{\text{p}} & (H + 2n)/2 \\ 0 & 0 & 1 \end{bmatrix}
$}
\label{K_p}
\end{equation}
here, \(f_x^{\text{p}} = \frac{W + 2n}{2 \tan\left(\frac{\text{FoVx}}{2}\right)}\) and \(f_y^{\text{p}} = \frac{H + 2n}{2 \tan\left(\frac{\text{FoVy}}{2}\right)}\) represent the updated focal lengths for the padded image.

\subsection{DETAILS OF OPTIMIZATION AND DENSIFICATION ALGORITHM
}
Our densify from refined depth algorithm is summarized in algorithm \ref{alg:densifyfromdepth}. Detailed densify of Adaptive Quantile-Based Geometric Consistency Densification algorithm is summarized in algorithm \ref{alg:qua}.

\begin{algorithm}
\small
    \caption{Densify from Depth \\
    K: camera intrinsic 3$\times$3\\
    E: Camera extrinsic \\
    M: Consistency Mask\\
    h: height\\
    w: weight\\
    N: normals\\
    I: images
    } \label{alg:densifyfromdepth}
    \begin{algorithmic}
        \State $M \gets \text{Extract } K, E \text{ from } V$ \Comment{$V$: Viewpoint Camera}
\State $y, x \gets \text{ meshgrid for coordinates}$ \Comment{Image coordinates}
\State $C_{3D} \gets K^{-1} \cdot \text{coord vectors}$ \Comment{Transform to 3D}
\State $D,I,N  \gets MaskFilter(M) $ \Comment{Filter data using $M$} 
\State $W_{3D} \gets C_{3D} \times D $ \Comment{$D$: Depth, $C$: cam2world}
\While{not converged} \Comment{Optimization loop}
    \State $F \gets \text{RGBtoSH}(I)$ \Comment{Spherical Harmonics}
    \State $S \gets ComputeDistance(W_{3D})$  \Comment{Scale}
    \State $R \gets \text{NormalstoRotation}(N)$ \Comment{Rotations}
    \State $O \gets InversSigmoid(W_{3D})$ \Comment{Opacity}
    
    \State $U \gets Init(W_{3D},F,S,R,O)$ \Comment{Initialization and Update}
\EndWhile
    \end{algorithmic}
\end{algorithm}

\begin{algorithm}
\small
    \caption{Adaptive Quantile-Based Geometric Consistency Densification \\
    K: camera intrinsic 3$\times$3\\
    E: Camera extrinsic \\
    id: Camera ID\\
    R: Gaussian render
    } \label{alg:qua}
    \begin{algorithmic}
        \State $ref_{id}, src_{id} \gets PairSelection(I,id)$ \Comment{Select pairs}
        \State $i \gets DensifyIteratons$ 
        \State $processed\_ids \gets Null$ \Comment{Every ref depth only densify once}
        \If {$isDepthDensfiyIterations(i)$} 
        \While{not converged}
        \If{$ref_{id}$ not in $processed\_ids$ }
            \For {$ref_{id}, src_{id}$ in $pairs$}
                \State $D_{ref}\gets R (ref_{id})  $ \Comment{Render ref Depths} 
                \State $M_{near}, M_{mid}, M_{far} \gets getQuantiles(D_{ref})$ \Comment{Get Masks}
                \For{$src_{id}$ in $src_{ids}$}
                    \State{$src_{cam} \gets getSrcCams(src_{id}, id)$}
                    \State $D_{src}\gets R (src_{cam})  $ \Comment{Render src Depths} 
                    \State $M_{near}\gets GeoCons (D_{ref}\cdot M_{near},  D_{src})$ \Comment{Get near Consistency mask}
                    \State $M_{mid}\gets GeoCons (D_{ref}\cdot M_{mid},  D_{src})$ \Comment{Get mid Consistency mask}
                    \State $M_{far}\gets GeoCons (D_{ref}\cdot M_{far},  D_{src})$ \Comment{Get far Consistency mask}
                \EndFor{}
            \State $M \gets M_{near}+ M_{mid}+M_{far}$ \Comment{Get combined mask}
            \State $G \gets DensifyfromDepth (D_{ref},M,K,S) $ \Comment{Densify from src Depth}
            \State $S \gets resetScale(G)$ \Comment{Reset scale}
            \State $O \gets resetOpacity(G)$ \Comment{Reset opacity}
            \State $processed\_ids \gets add (ref_{id})$ \Comment{Add processed ref image}
            \EndFor
            \Else { Skip Densify} \Comment{Skip densify if in $processed\_ids$}
            \EndIf
            
        \EndWhile
        
        \EndIf
    \end{algorithmic}
\end{algorithm}

\begin{algorithm}[!t]
\small
    \caption{Marching Cube with Multi-View Normal \\
    P: Point cloud\\
    K: camera intrinsic
    E: Camera extrinsic \\
    N: Normal\\
    D: Depth
    } \label{alg:mc}
    \begin{algorithmic}
\State $P' \gets crop(P) $ \Comment{Corp points by boundary}
\State $P' \gets filter (P')$ \Comment{{Remove floaters}}
\State $step \gets 0$
\For{$step$ in $STEPs$}
\For{$d,n $ in $D, N$} \Comment{Select multi view }
    \State $N_p' \gets getPointNormal(P')$ \Comment{$N_P'$: Point normal}
    \State $n_P' \gets W2C (N_P', K, E, d)$ \Comment{Point normal per view}
    \State $P'' \gets Soomth\&interp (N_p',n_P',K, E,)$ \Comment{Smooth and Interpolate points based on Normal}

\EndFor
\EndFor
\For{ $p$ in $P''$} \Comment{Split P'' into blocks}
    \State $d \gets getDnsity(p)$ \Comment{Get point density}
    \State $V \gets getVoxelSize(d,V_{min},V_{max})$ \Comment{Create Adaptive voxel size}
\EndFor
\If{$smooth$}
\State $V \gets smoothMesh(V)$ \Comment{Smooth voxel}
\EndIf
\State $v, t \gets MarchingCube(V)$ \Comment{v:vertices, t: triangles}
\State $Mesh \gets getMesh(v,t)$

    \end{algorithmic}
\end{algorithm}

\subsection{DETAILS OF MESH EXTRACTION ALGORITHM
}

Our mesh extraction method using Marching Cubes \cite{lorensen1998marching} is shown in algorithm \ref{alg:mc}.

\subsection{ADDITIONAL RESULTS}
\begin{figure}[!ht]
  \centering
   \includegraphics[width=\linewidth]{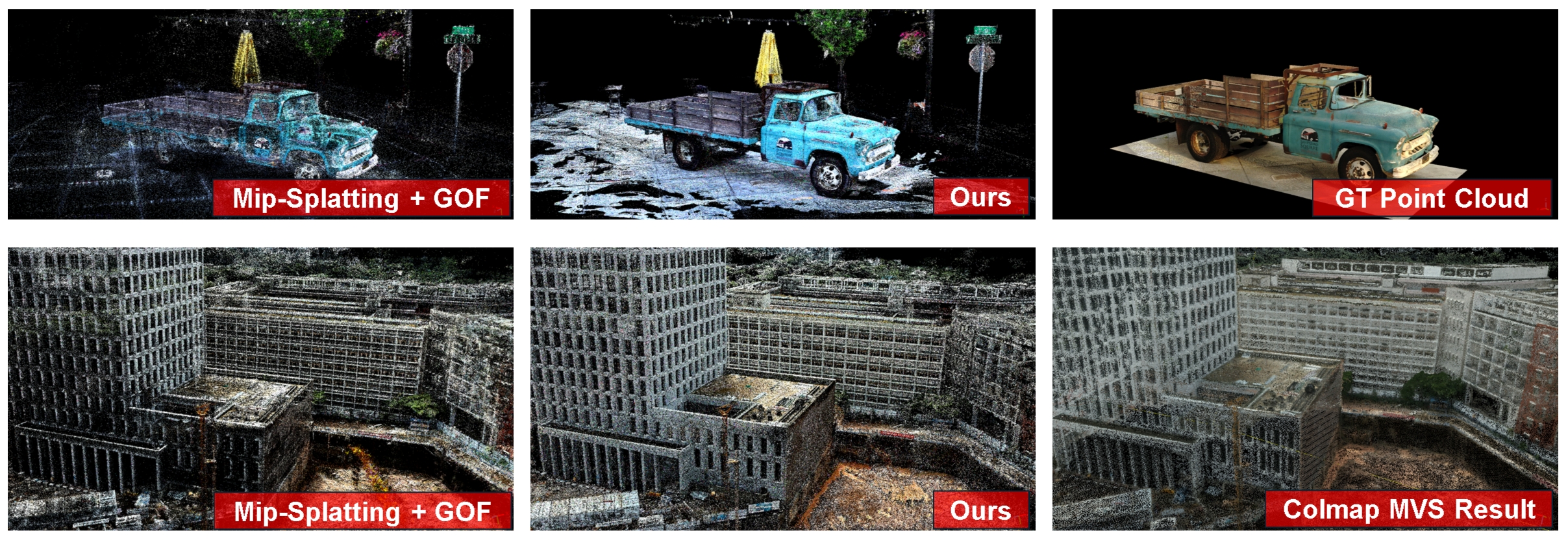}
   \caption{\textbf{Example of densification comparison.} We employ the GOF \cite{Yu2024GOF} densification method for comparison and provided reference point clouds for a thorough evaluation. From top to bottom: the \textit{Truck} from the Tanks and Temples dataset \cite{knapitsch2017tanks} and the \textit{Artsci} from the UrbanScene 3D dataset \cite{UrbanScene3D}. From left to right: Mip-Splatting \cite{yu2024mipsplatting} combined with GOF \cite{Yu2024GOF} densification strategy, our densification results, and the reference point clouds. It is evident that our densification approach provides a more uniform and dense Gaussian point cloud compared to the Mip-Splatting with GOF strategy.  }
   \label{fig:densification}
\end{figure}

\begin{figure}[!ht]
  \centering
   \includegraphics[width=\linewidth]{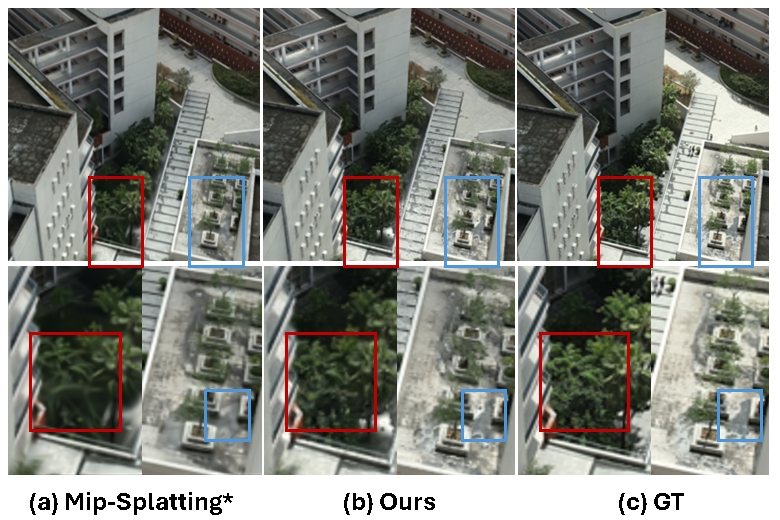}
   \caption{\textbf{Rendering results of different densification strategies.} We select the NVS rendering results from the \textit{Artsci} scene in the UrbanScene 3D dataset \cite{UrbanScene3D}. Our densification approach has demonstrated substantial capabilities in rendering complex geometric structures, such as trees (see red frame), effectively. Furthermore, it also excels in accurately rendering textures in areas with less geometric detail, such as the ground (see blue frame).  }
   \label{fig:us3d}
\end{figure}

Initially, Figure \ref{fig:densification} presents a example analysis of the the densification results in two distinct scenes \cite{knapitsch2017tanks, UrbanScene3D}. We selected Mip-Splatting \cite{yu2024mipsplatting} with GOF \cite{Yu2024GOF} densification as a compairsion to demonstrate that our method surpasses it in both uniformity and density of Gaussian point cloud data. Notably, a comparison was conducted against MVS point clouds generated by COLMAP \cite{schoenberger2016colmap} on the UrbanScene 3D dataset \cite{UrbanScene3D}. To ensure fairness, all three methods were evaluated using a resolution of 512. The results indicate that our method significantly outperforms the GOF \cite{Yu2024GOF} densification approach utilized in Mip-Splatting \cite{yu2024mipsplatting}, particularly in low-resolution settings on large-scale datasets. The advantage of our densification technique lies in its ability to enhance the overall rendering quality (as shown in Figure \ref{fig:us3d}) while providing superior mesh extraction results.

\begin{figure*}[!ht]
  \centering
   \includegraphics[width=\linewidth]{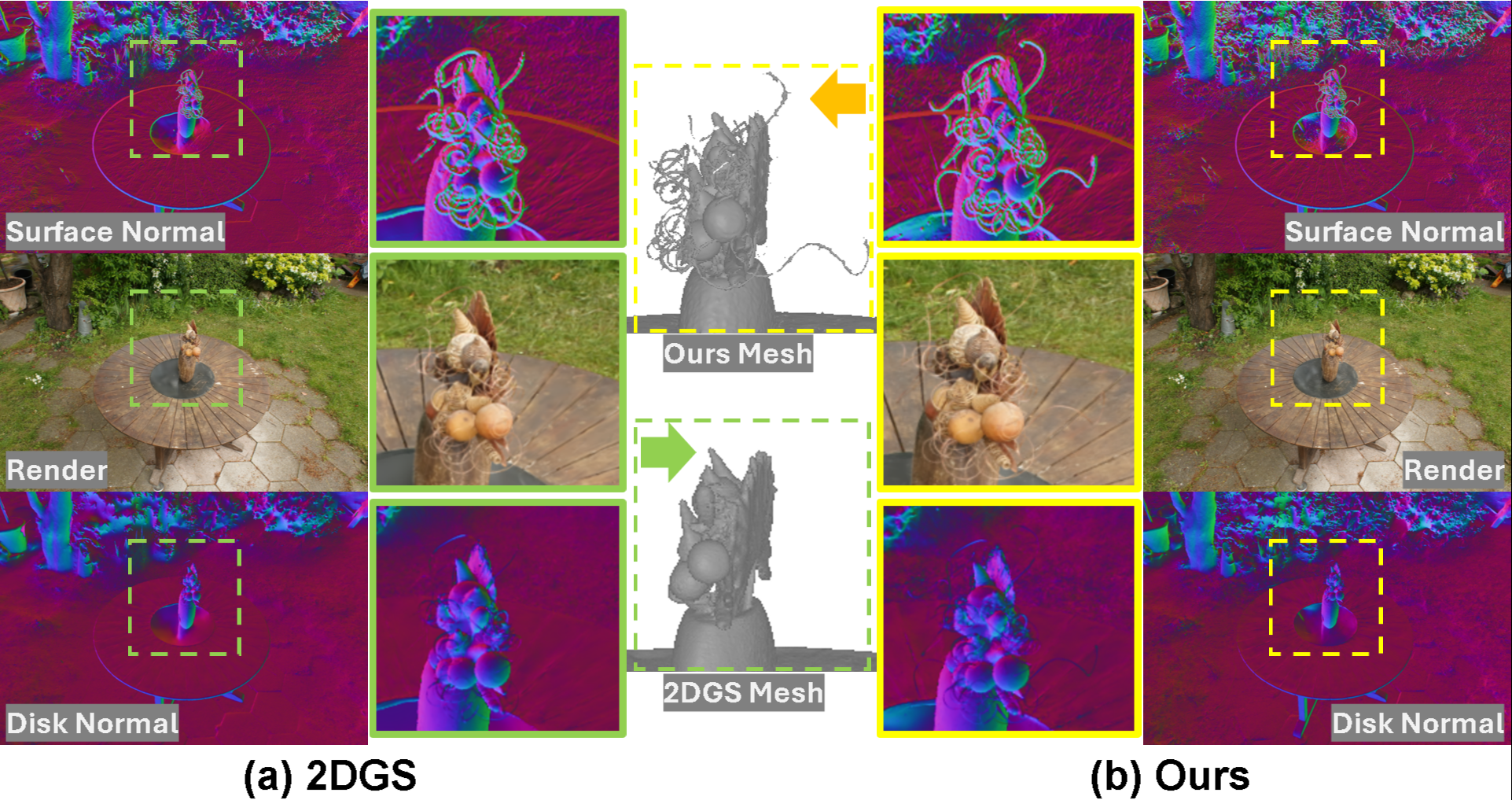}
   \caption{\textbf{Example of full comparison.} Comparison of NVS, mesh, depth normal, and disk normal results on the Mip NeRF 360 ]\cite{barron2022mipnerf360} \textit{garden }dataset. }
   \label{fig:mesh_depth_normal}
\end{figure*}
Figure \ref{fig:mesh_depth_normal} illustrates a comprehensive comparison including rendered scenes, surface normals, disk normals, and the extracted meshes. It is evident from the comparison that our method surpasses 2DGS \cite{huang20242dgs} in rendering more complete normals and, by integrating densified point clouds with the Marching Cubes  algorithm \ref{alg:mc}, produces more detailed meshes. 

\begin{figure*}[!ht]
  \centering
   \includegraphics[width=\linewidth]{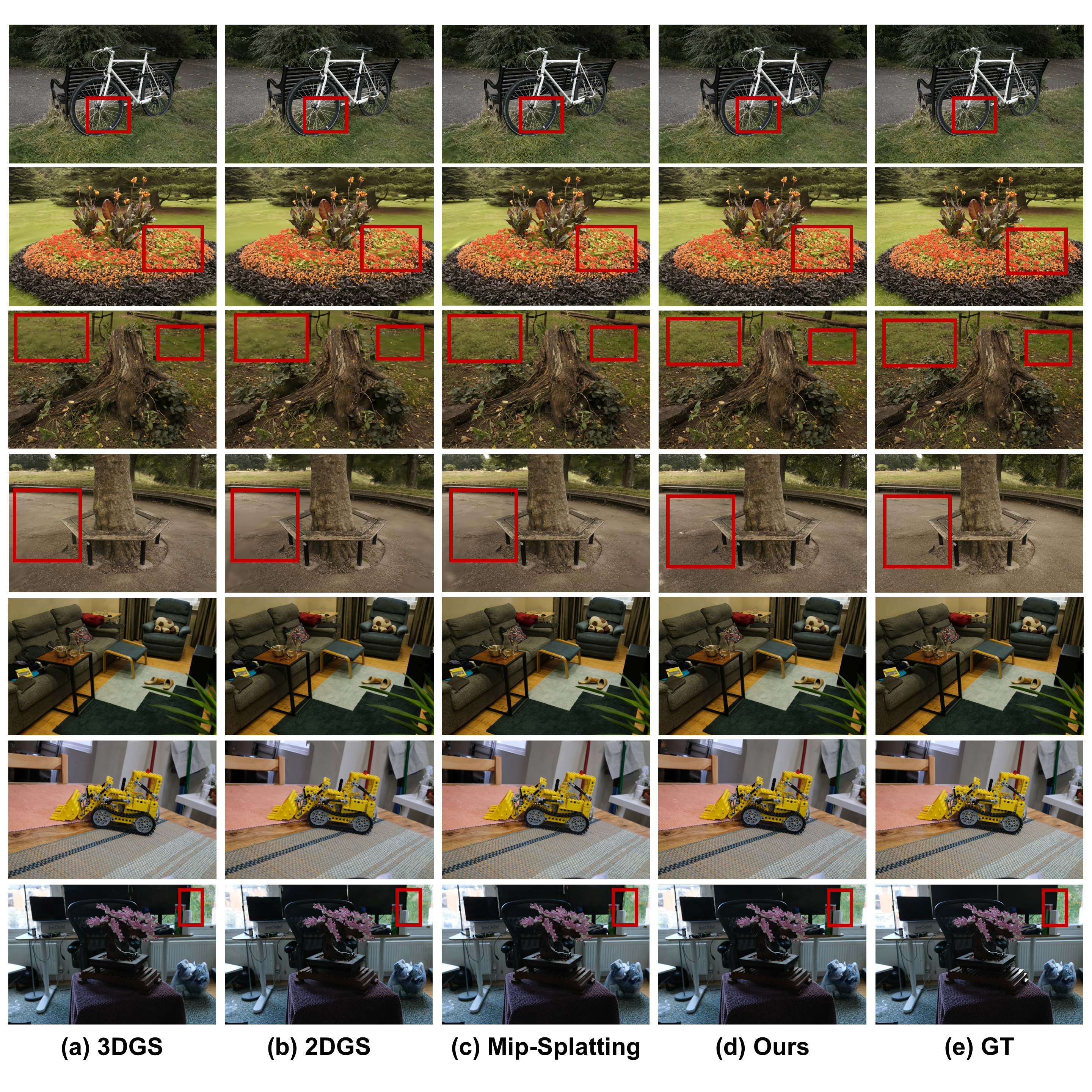}
   \caption{\textbf{Novel View Synthesis Comparison.} Comparison of Per View NVS results on the Mip NeRF 360 ]\cite{barron2022mipnerf360}. }
   \label{fig:360full}
\end{figure*}

Table \ref{perscene} provides a detailed metric comparison for each scene in the Mip-NeRF 360 dataset \cite{barron2022mipnerf360}. And Figure \ref{fig:360full} provides a detailed comparison of each scene in the Mip-NeRF 360 dataset \cite{barron2022mipnerf360}.

\begin{figure*}[!ht]
  \centering
   \includegraphics[width=\linewidth]{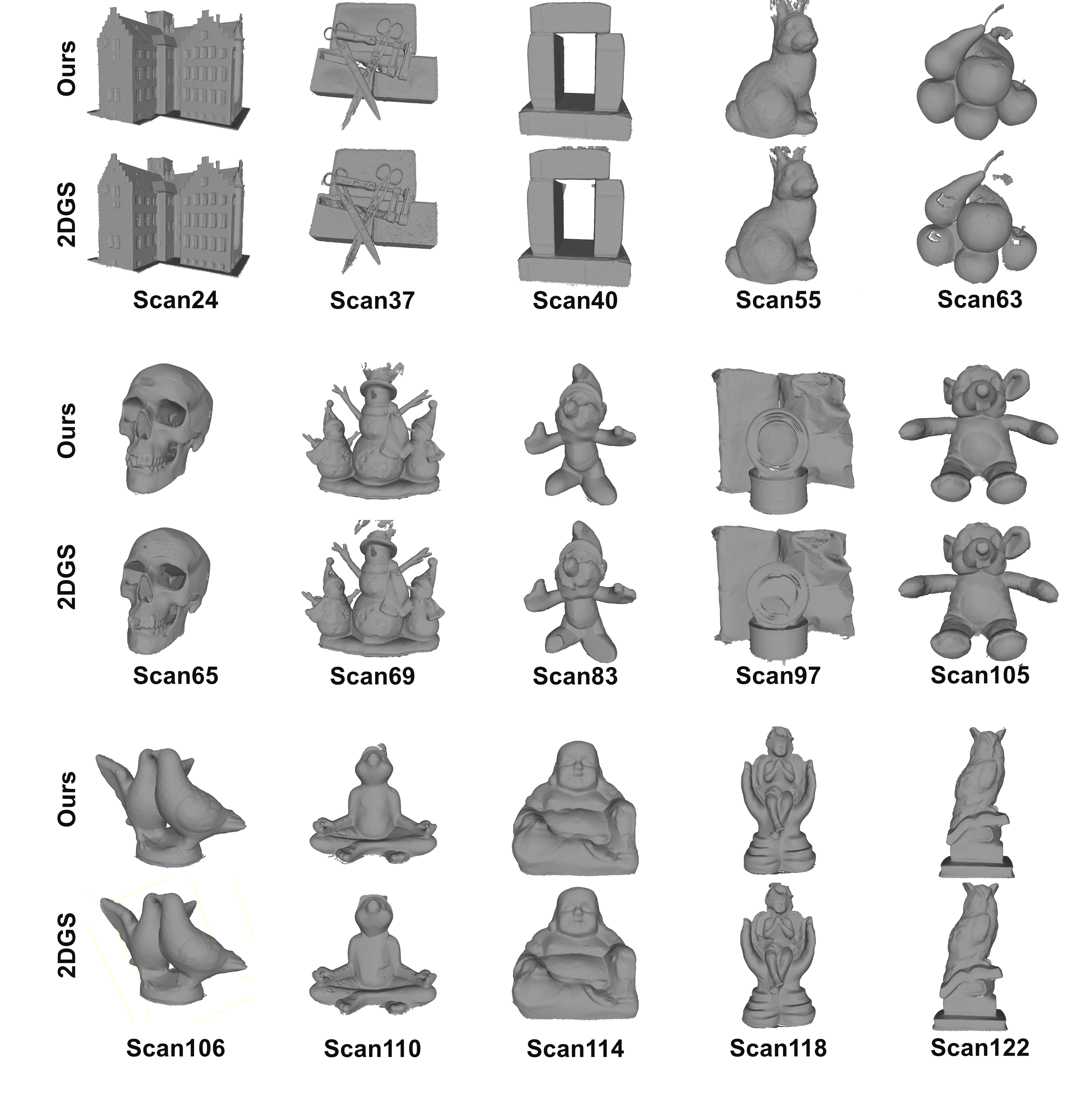}
   \caption{\textbf{Mesh Comparison.} Comparison of Mesh results on the DTE datasets \cite{jensen2014dtu}.}
   \label{fig:dtu}
\end{figure*}

Additionally, Figure \ref{fig:dtu} showcases a comparison of mesh extraction on the DTU dataset.
\begin{table*}[ht]
\centering
\small

\begin{tabular}{l|lllll|llll}
              \multicolumn{10}{c}{\textbf{PSNR}}                                                 \\
              
              & \textit{bicycle} & \textit{flowers }& \textit{garden} & \textit{stump} & \textit{treehill} & \textit{room} & \textit{counter} & \textit{kitchen} & \textit{bonsai} \\
              \hline
NeRF          & 21.76 & 19.40& 23.11& 21.73& 21.28& 28.56& 25.67& 26.31& 26.81
\\
Instant NGP   & 22.79 & 19.19& 25.26& 24.80& 22.46& 30.31& 26.21& 29.00& 31.08\\
Mip-NeRF 360  & 24.40 & 21.64& 26.94& 26.36& 22.81& 31.40& \cellcolor{red!30}29.44& \cellcolor{red!30}32.02 &\cellcolor{red!30}33.11\\
3DGS          &25.25 & 21.52& 27.41& 26.55& 22.49& 30.63& 28.70& 30.32& 31.98
\\
2DGS          & 25.09 &21.19& 27.14& 26.59& 22.69& 31.16& 28.63& 30.82& 31.81\\
Mip-Splatting &\cellcolor{yellow!30}25.72 &\cellcolor{yellow!30}21.93 &\cellcolor{yellow!30}27.76 &\cellcolor{yellow!30}26.94& \cellcolor{orange!30}22.98&\cellcolor{orange!30} 31.74& 29.16 &31.55& 32.31 \\

Mip-Splatting* &\cellcolor{red!30}25.81 &\cellcolor{orange!30}21.97&\cellcolor{red!30} 27.86&\cellcolor{red!30}27.01& \cellcolor{red!30}23.06&\cellcolor{red!30} 31.80& \cellcolor{yellow!30}29.26 &\cellcolor{yellow!30}31.66& \cellcolor{yellow!30}32.53 \\
\hline
Ours          & \cellcolor{orange!30}25.76 &  \cellcolor{red!30}21.98  & \cellcolor{orange!30}27.78 & \cellcolor{orange!30}26.98 & \cellcolor{yellow!30}22.93 &\cellcolor{yellow!30} 31.60 &\cellcolor{orange!30} 29.36 & \cellcolor{orange!30}31.74& \cellcolor{orange!30}32.60   \\

                \multicolumn{10}{c}{}\\
              \multicolumn{10}{c}{\textbf{SSIM}}                                                 \\
              & \textit{bicycle} & \textit{flowers }& \textit{garden} & \textit{stump} & \textit{treehill} & \textit{room} & \textit{counter} & \textit{kitchen} & \textit{bonsai} \\
              \hline
NeRF& 0.455& 0.376& 0.546& 0.453& 0.459& 0.843& 0.775& 0.749& 0.792
\\
Instant NGP   &0.540& 0.378& 0.709& 0.654& 0.547& 0.893& 0.845& 0.857& 0.924       \\
Mip-NeRF 360  & 0.693 &0.583& 0.816& 0.746& 0.632& 0.913 &0.895 &0.920 &0.939      \\
3DGS          &0.771& 0.605& 0.868& 0.775& 0.638& 0.914& 0.905& 0.922& 0.938
\\
2DGS          &0.742& 0.584& 0.857& 0.777& 0.641& \cellcolor{yellow!30}0.933&\cellcolor{yellow!30} 0.920 &\cellcolor{yellow!30}0.940& \cellcolor{yellow!30}0.948\\
Mip-Splatting &\cellcolor{yellow!30}0.780 &\cellcolor{yellow!30}0.623 &\cellcolor{yellow!30}0.875& \cellcolor{yellow!30}0.786&\cellcolor{yellow!30}0.655& 0.928& 0.916& 0.933 & \cellcolor{yellow!30}0.948  \\

Mip-Splatting* &\cellcolor{orange!30}0.788 &\cellcolor{red!30}0.632 &\cellcolor{red!30}0.880& \cellcolor{red!30}0.791&\cellcolor{red!30}0.660& \cellcolor{orange!30}0.938& \cellcolor{orange!30}0.926&\cellcolor{orange!30} 0.947 & \cellcolor{red!30}0.960  \\\hline

Ours          & \cellcolor{red!30} 0.791      &  \cellcolor{orange!30} 0.631     &   \cellcolor{orange!30} 0.871    & \cellcolor{orange!30} 0.789     &   \cellcolor{yellow!30}0.652       & \cellcolor{red!30}  0.939   &    \cellcolor{red!30}0.927     & \cellcolor{red!30}   0.948     &     \cellcolor{orange!30}0.958   \\

\multicolumn{10}{c}{}\\
              \multicolumn{10}{c}{\textbf{LPIPS}}                                                \\
              & \textit{bicycle} & \textit{flowers }& \textit{garden} & \textit{stump} & \textit{treehill} & \textit{room} & \textit{counter} & \textit{kitchen} & \textit{bonsai} \\
              \hline
NeRF          & 0.536 & 0.529 & 0.415 & 0.551 & 0.546 & 0.353 & 0.394 & 0.335 & 0.398 
       \\
Instant NGP   &0.398& 0.441 &0.255& 0.339 &0.420& 0.242& 0.255& 0.170& 0.198
       \\
Mip-NeRF 360  &0.289& 0.345& 0.164& 0.254& 0.338& 0.211& 0.203& 0.126& 0.177\\
3DGS          &\cellcolor{yellow!30}0.205& 0.336&\cellcolor{yellow!30} 0.103 & \cellcolor{yellow!30}0.210 &\cellcolor{yellow!30}0.317 &0.220 &0.204 &0.129 &0.205       \\
2DGS          & 0.209& 0.348& 0.111& 0.219 &0.333& 0.224& 0.207& 0.133 &0.204       \\
Mip-Splatting &0.206& \cellcolor{yellow!30}0.331& \cellcolor{yellow!30}0.103& \cellcolor{yellow!30}0.209& 0.320& \cellcolor{yellow!30}0.192&\cellcolor{yellow!30} 0.179 &\cellcolor{yellow!30}0.113&\cellcolor{yellow!30} 0.173       \\
Mip-Splatting & \cellcolor{orange!30}0.193&\cellcolor{orange!30} 0.329& \cellcolor{red!30}0.102&\cellcolor{red!30} 0.207&\cellcolor{red!30} 0.313& \cellcolor{orange!30}0.190& \cellcolor{orange!30}0.172 &\cellcolor{orange!30}0.109&\cellcolor{orange!30} 0.165       \\ \hline

Ours          & \cellcolor{red!30} 0.186       & \cellcolor{red!30}  0.322      &  \cellcolor{orange!30}0.101      & 0.213     &  \cellcolor{orange!30}0.316        & \cellcolor{red!30} 0.188     &   \cellcolor{red!30} 0.169     &  \cellcolor{red!30}   0.106    & \cellcolor{red!30}0.161       \\

\end{tabular}

\caption{\textbf{Per Scene Novel View Synthesis Results.} We present the results of NVS for each scene on the Mip-NeRF 360 dataset \cite{barron2022mipnerf360}.  Mip-Splatting* means we report Mip-Splatting \cite{yu2024mipsplatting} metrics with GOF \cite{Yu2024GOF} densification.}
\label{perscene}
\end{table*}

\begin{figure*}[!ht]
  \centering
   \includegraphics[width=\linewidth]{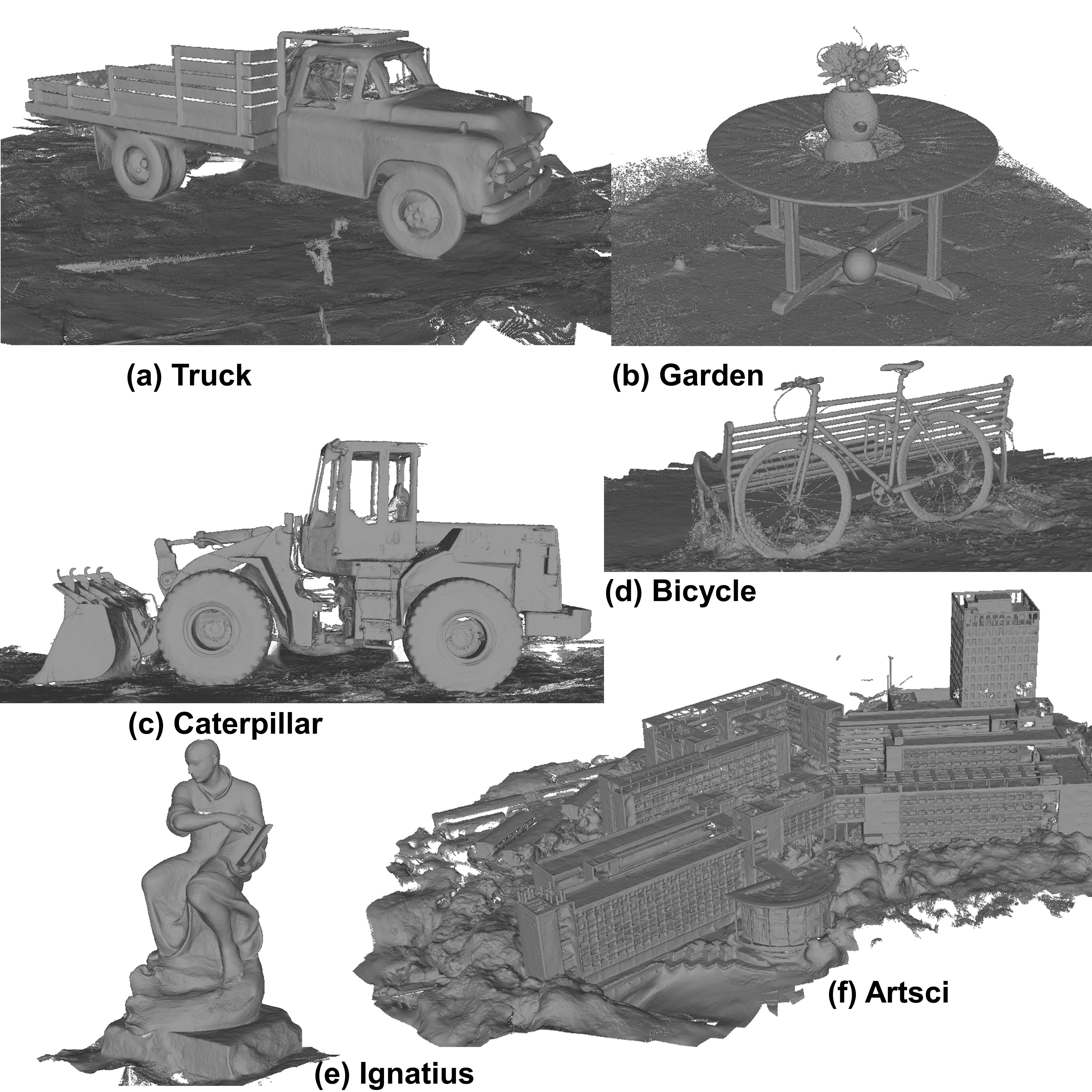}
   \caption{\textbf{More Mesh Results.} Ours Mesh results on Tanks and Temples \cite{knapitsch2017tanks}, Mip-NeRF 360 \cite{barron2022mipnerf360}, and UrbanScene 3D \cite{UrbanScene3D} datasets.}
   \label{fig:more_mesh}
\end{figure*}

\begin{figure*}[!ht]
  \centering
   \includegraphics[width=\linewidth]{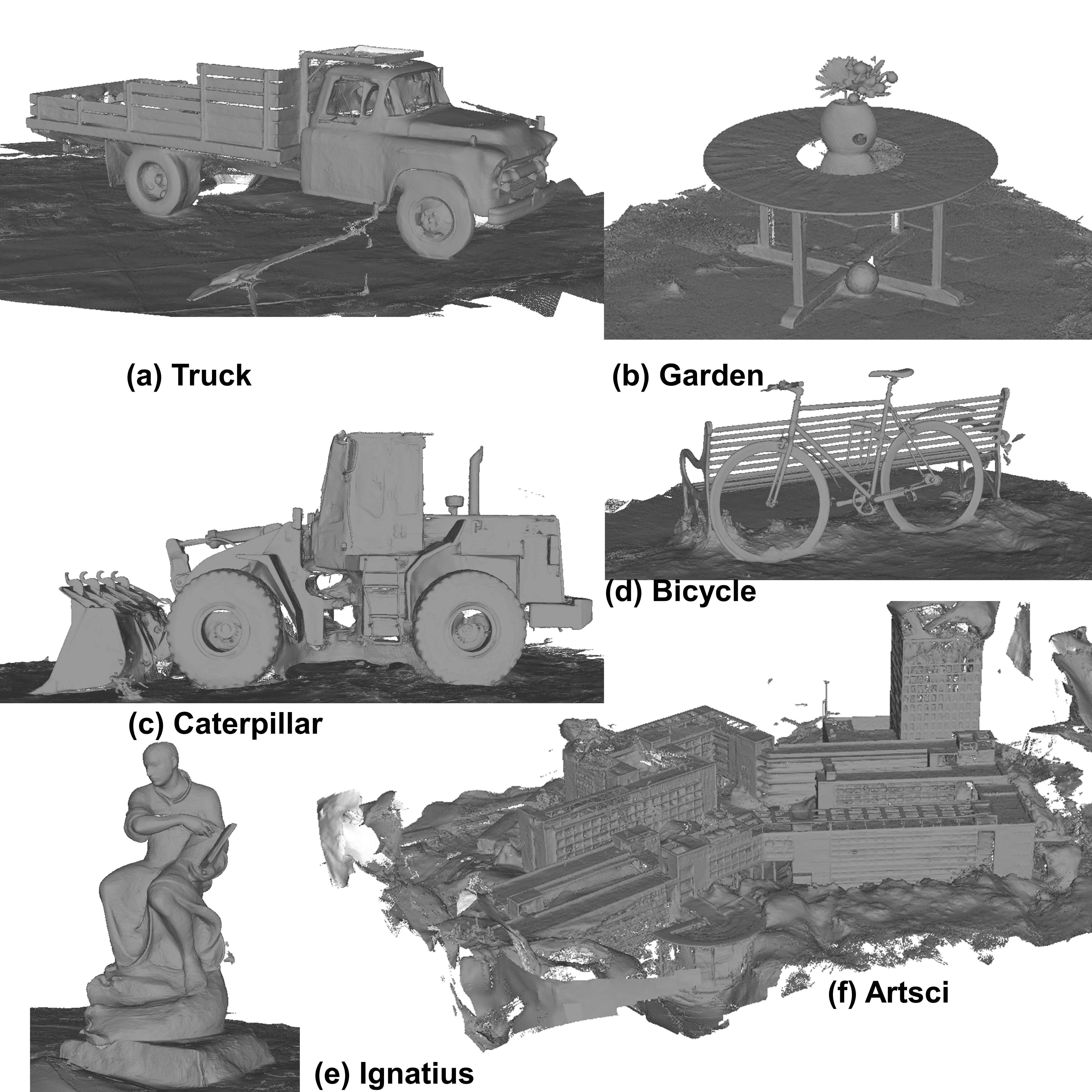}
   \caption{\textbf{More Mesh Results.} 2DGS Mesh results on Tanks and Temples \cite{knapitsch2017tanks}, Mip-NeRF 360 \cite{barron2022mipnerf360}, and UrbanScene 3D \cite{UrbanScene3D} datasets.}
   \label{fig:2d_gs_more_mesh}
\end{figure*}


\end{document}